%% file: main.tex
\documentclass[10pt,twocolumn,letterpaper]{article}

\usepackage[pagenumbers]{cvpr} 

\input{preamble}

\definecolor{cvprblue}{rgb}{0.21,0.49,0.74}
\usepackage[pagebackref,breaklinks,colorlinks,citecolor=cvprblue]{hyperref}

\usepackage{etoolbox}
\makeatletter
\patchcmd\Hy@backout{\@auxout}{\@mainaux}{}{\fail}
\patchcmd\Hy@backout{\@auxout}{\@mainaux}{}{\fail} 
\makeatother

\usepackage[capitalize]{cleveref}
\crefname{section}{Sec.}{Secs.}
\crefname{section}{Section}{Sections}
\crefname{table}{Table}{Tables}
\crefname{table}{Tab.}{Tabs.}

\def\paperID{14090} 
\def\confName{CVPR}
\def\confYear{2024}

\title{EGTR: Extracting Graph from Transformer for Scene Graph Generation}

\author{Jinbae Im\textsuperscript{\rm 1}
    \quad JeongYeon Nam\textsuperscript{\rm 1}
    \quad Nokyung Park\textsuperscript{\rm 1,2,3}\thanks{~Most work was done during the internship at NAVER Cloud AI.}
    \quad Hyungmin Lee\textsuperscript{\rm 2}
    \quad Seunghyun Park\textsuperscript{\rm 1} \\
    \textsuperscript{\rm 1}NAVER Cloud AI
    \quad \textsuperscript{\rm 2}NAVER
    \quad \textsuperscript{\rm 3}Korea University\\
    \{jinbae.im, jy.nam, nokyung.park99, hyungmin.lee, seung.park\}@navercorp.com
}

\begin{document}
\maketitle

\begin{abstract}
\input{tex/0_abstract.tex}
\end{abstract}

\section{Introduction}\label{sec:intro}
\input{tex/1_introduction.tex}
\section{Related Work}
\input{tex/2_relatedwork.tex}
\section{Method}
\input{tex/3_method.tex}
\section{Experiments}
\input{tex/4_experiment.tex}
\section{Conclusion}
\input{tex/5_conclusion.tex}

\clearpage

\section*{Authors' Contribution}
\input{tex/7_contribution.tex}

{
    \small
    \bibliographystyle{ieeenat_fullname}
    \bibliography{main}
}



\clearpage
\setcounter{section}{0}
\setcounter{table}{0}
\setcounter{figure}{0}
\maketitlesupplementary
\input{tex/6_supplementary.tex}


\end{document}

%% file: preamble.tex
\usepackage[dvipsnames]{xcolor}

\newcommand{\myparagraph}[1]{\vspace{4pt}\noindent{\bf #1}}
\usepackage{graphicx}
\usepackage{amsmath}
\usepackage{amssymb}
\usepackage{booktabs}
\usepackage{bbm}
\usepackage[accsupp]{axessibility}


%% file: tex/0_abstract.tex
Scene Graph Generation (SGG) is a challenging task of detecting objects and predicting relationships between objects. After DETR was developed, one-stage SGG models based on a one-stage object detector have been actively studied. However, complex modeling is used to predict the relationship between objects, and the inherent relationship between object queries learned in the multi-head self-attention of the object detector has been neglected. We propose a lightweight one-stage SGG model that extracts the relation graph from the various relationships learned in the multi-head self-attention layers of the DETR decoder. By fully utilizing the self-attention by-products, the relation graph can be extracted effectively with a shallow relation extraction head. Considering the dependency of the relation extraction task on the object detection task, we propose a novel relation smoothing technique that adjusts the relation label adaptively according to the quality of the detected objects. By the relation smoothing, the model is trained according to the continuous curriculum that focuses on object detection task at the beginning of training and performs multi-task learning as the object detection performance gradually improves. Furthermore, we propose a connectivity prediction task that predicts whether a relation exists between object pairs as an auxiliary task of the relation extraction. We demonstrate the effectiveness and efficiency of our method for the Visual Genome and Open Image V6 datasets. Our code is publicly available at~\url{https://github.com/naver-ai/egtr}.

%% file: tex/1_introduction.tex
\begin{figure}
    \centering
    \begin{minipage}{.3\textwidth}
        \begin{subfigure}[t]{\linewidth}
            \includegraphics[width=\textwidth]{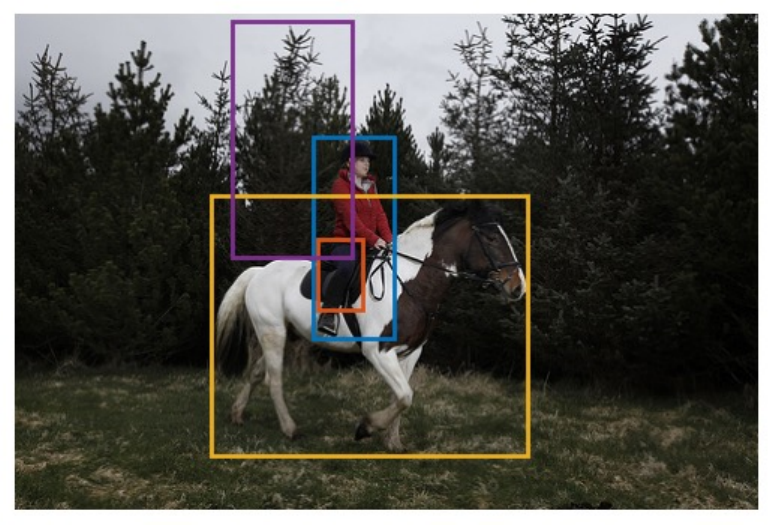}
            \caption{Visual Genome example}
            \label{fig:intro1}
        \end{subfigure}
    \end{minipage}
    \quad
    \begin{minipage}{.3\linewidth}
        \begin{subfigure}[t]{\linewidth}
            \includegraphics[width=\textwidth]{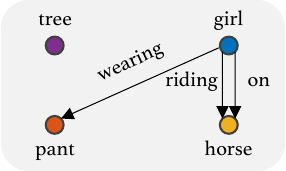}
            \caption{Scene Graph}
            \label{fig:intro2}
        \end{subfigure} \\
        \begin{subfigure}[b]{\linewidth}
            \vspace{1.5mm}
            \includegraphics[width=\textwidth]{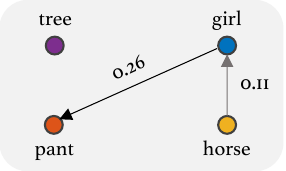}
            \caption{Attention Graph}
            \label{fig:intro3}
        \end{subfigure}
    \end{minipage}
    \vspace{-.5em}
\caption{\textbf{Motivation.} SGG task aims to predict scene graph (\cref{fig:intro2}) with objects as nodes and relations as edges. We draw a plausible attention graph (\cref{fig:intro3}) simply by connecting objects with high attention weights to edges from the self-attention layers of the pre-trained DETR.
It shows the potential for the self-attention from the object detector to contain rich information that aids in predicting the relations of the scene graph.
}
\label{fig:intro}
\vspace{-4mm}
\end{figure}

Scene Graph Generation (SGG)~\cite{johnson2015image} aims to generate a scene graph that represents objects as nodes and relationships between objects as edges from an image as shown in~\cref{fig:intro2}. SGG is a challenging task as it is required to go beyond simply detecting objects but predicting the relationships between them based on a comprehensive understanding of the scene. Since a scene graph provides structural information of the image, it can be used for various vision tasks that require a higher level of understanding and reasoning about images such as image captioning~\cite{gao2018image,kim2019dense,yang2019auto}, image retrieval~\cite{johnson2015image,schuster2015generating}, and visual question answering~\cite{li2019relation,zhang2019empirical}.
 \begin{figure*}
    \centering
    \begin{minipage}{.31\textwidth}
        \begin{subfigure}[t]{\linewidth}
            \includegraphics[width=\textwidth]{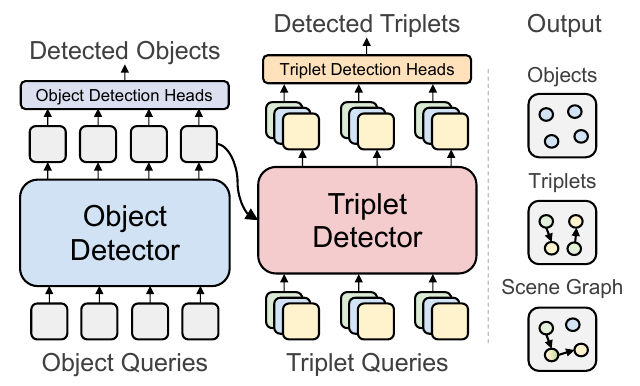}
            \caption{Object-Triplet Detection Models}
            \label{fig:fig_one_stage_1}
        \end{subfigure}
    \end{minipage}
    \begin{minipage}{.31\textwidth}
        \begin{subfigure}[t]{\linewidth}
            \includegraphics[width=\textwidth]{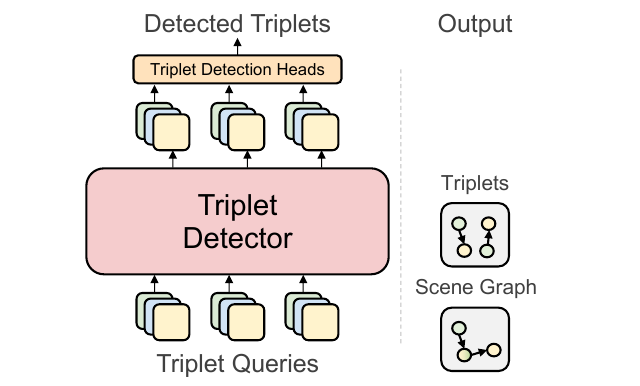}
            \caption{Triplet Detection Models}
            \label{fig:fig_one_stage_2}
        \end{subfigure}
    \end{minipage}
    \begin{minipage}{.31\textwidth}
        \begin{subfigure}[t]{\linewidth}
            \includegraphics[width=\textwidth]{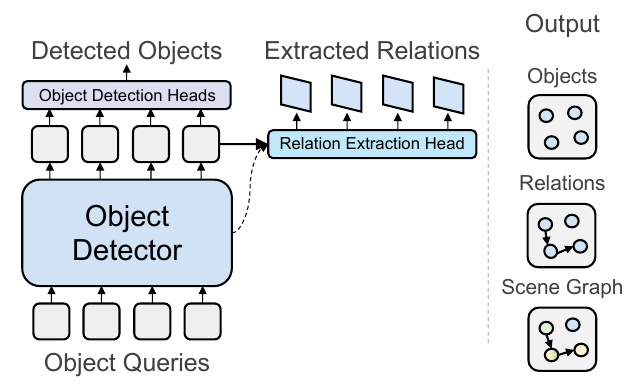}
            \caption{Relation Extraction Models}
            \label{fig:fig_one_stage_3}
        \end{subfigure}
    \end{minipage}
    \vspace{-.5em}
\caption{\textbf{Comparison with existing one-stage SGG models.} 
(a) Object-Triplet Detection Models introduce additional triplet queries and a triplet detector to the object detector. The triplet detector requires additional modules to incorporate information from the object detector into the triplet queries. 
(b) Triplet Detection Models focus on detecting triplets directly without an object detector. Objects without relations may not be detected. 
(c) Relation Extraction Models extract relations from the object detector without a separate triplet detector. In particular, ours extracts relations more effectively by utilizing by-products from the self-attention of the object detector.
}
\label{fig:one_stage_models}
\vspace{-4mm}
\end{figure*}

Most previous studies~\cite{zellers2018neural,xu2017scene,tang2019learning,lin2020gps,li2021bipartite} took two-stage SGG approaches that detect objects first and then predict their relations. 
However, these approaches incurred high computational costs and risks of error propagation from the object detection.
To address the drawbacks of the two-stage approaches, one-stage SGG models that perform object detection and relation prediction at once~\cite{cong2023reltr,li2022sgtr,liu2021fully,newell2017pixels,shit2022relationformer,teng2022structured,khandelwal2022iterative} have been studied recently, leveraging one-stage object detectors such as DETR~\cite{carion2020end}.
Since edges in a scene graph can be represented as subject-predicate-object triplets, many studies embraced triplet-based approaches as shown in~\cref{fig:fig_one_stage_1,fig:fig_one_stage_2}. However, object-triplet detection models~\cite{cong2023reltr,li2022sgtr} required sophisticated triplet detectors to obtain necessary information for triplet queries from the object detector. Moreover, triplet detection models~\cite{khandelwal2022iterative,teng2022structured} lacked the ability to detect objects without relation such as ``tree" in~\cref{fig:intro2}, by focusing solely on the triplet detection without an object detector. Considering that objects without relations account for more than $42\%$ in Visual Genome~\cite{krishna2017visual} data, their priority lies in detecting a sub-graph rather than the complete scene graph.

To address the shortcomings of the existing one-stage SGG models, we focus on the relationships between objects inherent in the object detector.
As shown in~\cref{fig:intro1}, objects are related to each other. For instance, when a horse appears in a scene, a person is likely to appear in the scene, and clothing such as hats, jackets, and pants often depicts the person's current situation.
From this intuition, there has been a long belief~\cite{hu2018relation,liu2018structure,chen2018context} that modeling relationship or context between objects would be beneficial to object detection task.
Accordingly, recent one-stage object detectors~\cite{carion2020end,zhu2020deformable,sun2021sparse} have incorporated self-attention layers~\cite{vaswani2017attention} to implicitly model the relationships among the object queries. 
We hypothesize that self-attention between object queries learned in a one-stage object detector might contain valuable information for predicting triplet outputs. In our preliminary investigation, we are able to extract a plausible attention graph by simply connecting two object queries with high attention weights from the pre-trained DETR, as shown in~\cref{fig:intro3}. It shows the potential that the attention weights between object queries can be interpreted as relations between them.

From the findings, we propose a lightweight one-stage scene graph generator \textbf{EGTR}, which stands for \textbf{E}xtracting \textbf{G}raph from \textbf{TR}ansformer. 
We design the model to comprehensively leverage the by-products of the object detector, eliminating the need for a separate triplet detector, as depicted in~\cref{fig:fig_one_stage_3}. 
From the multi-head self-attention layers of the object detector, we regard an attention query and key, where their relations are learned in the attention weights, as the subject entity and object entity, respectively. 
Subsequently, we leverage a shallow classifier to predict relations between them.
Due to the abundant information about the relationships among the objects present in the by-products derived from all layers of self-attention, we can effectively extract the scene graph. 

Since the relation extraction task is dependent on the object detection task, we speculate that performing relation extraction without sufficiently learned representations of the object queries might be harmful.
Therefore, we devise a novel adaptive smoothing technique that smooths the value of the ground truth relation label based on the object detection performance.
With the adaptive smoothing,
the model is trained with a continuous curriculum that initially focuses on object detection and gradually performs multi-task learning.
Furthermore, we propose a connectivity prediction task as an auxiliary task for relation extraction, aiming to predict the existence of any relationship between a subject entity and an object entity. This auxiliary task facilitates the acquisition of representations for the relation extraction.

To verify the effectiveness of the proposed method, we conduct experiments on two representative SGG datasets: Visual Genome~\cite{krishna2017visual} and Open Images V6~\cite{kuznetsova2020open}.
By actively utilizing the by-products of the object detector, EGTR shows the best object detection performance and comparable triplet detection performance, with the fewest parameters and the fastest inference speed.

Our main contributions can be summarized as follows: 
\begin{itemize}
\item We propose EGTR that generates scene graphs efficiently and effectively by utilizing the multi-head self-attention by-products derived from the object detector.
\item We present adaptive smoothing, enabling effective multi-task learning for both object detection and relation extraction. In addition, the proposed connectivity prediction offers clues to the relation extraction.
\item Our comprehensive experiments show the superiority of the proposed model framework and the effectiveness of the devised training techniques.
\end{itemize}

%% file: tex/2_relatedwork.tex
SGG models can be categorized into two groups: two-stage models and one-stage models. 
For two-stage models~\cite{lu2016visual,xu2017scene,woo2018linknet,zellers2018neural,lin2020gps,koner2020relation,li2021bipartite,dhingra2021bgt,tang2019learning,lu2021context,zheng2022learning,sudhakaran2023vision,li2023compositional,min2023environment}, separate object detection model and relation prediction model are trained sequentially.
They usually detect $N$ objects from the off-the-shelf object detector such as Faster R-CNN~\cite{ren2015faster}, and then all possible combinations of the detected objects are fed into the relation prediction model to predict the relations between each object pair. Although they showed high relation extraction performance, 
they have the inherent limitation that the object detector that helps with relation extraction is trained separately, resulting in a significant increase in model complexity.

As for the one-stage models~\cite{liu2021fully,newell2017pixels,cong2023reltr,li2022sgtr,khandelwal2022iterative,teng2022structured,shit2022relationformer}, the object detection and relation prediction are trained in an end-to-end manner.
Early studies~\cite{liu2021fully,newell2017pixels} proposed fully convolutional SGG models and took pixel-based approaches.
After DETR~\cite{carion2020end} brought a huge success as a Transformer~\cite{vaswani2017attention}-based one-stage object detector, many one-stage SGG studies are based on one-stage object detectors~\cite{carion2020end,zhu2020deformable, sun2021sparse}. They efficiently modeled SGG by introducing object queries or triplet queries. We categorize them into three distinct groups: (a) object-triplet detection models, (b) triplet prediction models, and (c) relation extraction models, as shown in~\cref{fig:one_stage_models}. 

\vspace{4pt}
\noindent
\textbf{Object-Triplet Detection Models.} 
Object-triplet detection models are characterized by introducing additional triplet queries and building a triplet predictor on top of the object detector as shown in~\cref{fig:fig_one_stage_1}. RelTR~\cite{cong2023reltr} introduced paired subject queries and object queries, and SGTR~\cite{li2022sgtr} introduced compositional queries decoupled into subjects, objects, and predicates. As the introduced queries are initialized without prior cues from the object detector, the triplet predictor requires modules to incorporate information from the outputs of the object detector and modules for triplet queries to exchange information with each other. It led to the triplet predictor having an intricate structure. 
In contrast, we refrain from introducing additional queries and regard the attention queries and attention keys, whose relationships are learned in the object detector, as subject and object queries, respectively.

\vspace{4pt}
\noindent
\textbf{Triplet Detection Models.}
Triplet detection models directly detect triplets using triplet queries without an object detector, as shown in~\cref{fig:fig_one_stage_2}. Iterative SGG~\cite{khandelwal2022iterative} introduced subject, object, and predicate queries and modeled the conditional dependencies among them with separate subject, object, and predicate multi-layer Transformer decoders. 
Inspired by Sparse R-CNN~\cite{sun2021sparse}, Structured Sparse R-CNN (SSR-CNN)~\cite{teng2022structured} designed triplet queries consisting of subject box, object box, subject, object, and predicate queries. 
Since it was difficult to train the model only with sparse triplet annotations, SSR-CNN introduced somewhat intricate training details: it detected object pairs with a Siamese Sparse RCNN model and auxiliary queries and performed additional triplet matching using the detected object pairs as a pseudo label.
Although they show excellent triplet detection performance, 
the model architecture became even more intricate to detect subjects and objects separately since an explicit object detector is not used.
Last but not least, the triplet prediction models focus on detecting a sub-graph composed only of objects with relations, overlooking objects in an image that lack explicit relations.

\vspace{4pt}
\noindent
\textbf{Relation Extraction Models.}
Relation extraction models extract a scene graph using a lightweight relation predictor without separate triplet queries or triplet detector. 
Relationformer~\cite{shit2022relationformer} added a special ``[rln]" token to capture global information in conjunction with object queries. They concatenated the final hidden representations of the object query pairs and the relation token, followed by a shallow fully connected network for relation prediction.
In this work, in conjunction with the final hidden representations, we use the inherent relationship information among the object queries learned in the multi-head self-attention layer of the object detector. Furthermore, we propose training techniques to boost multi-task learning, leading to significantly improved performance with architectural simplicity.

%% file: tex/3_method.tex
\begin{figure*}[t!]
    \begin{center}
        \includegraphics[width=1.0\linewidth]{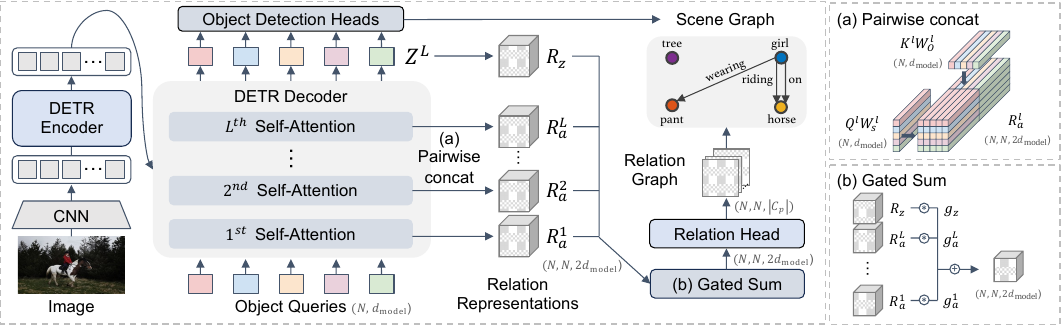}
    \end{center}
    \vspace{-.5em}
    \caption{\textbf{The overall architecture of EGTR}. 
    We present a novel lightweight relation extractor, EGTR, which fully utilizes the self-attention of the DETR decoder. We extract query and key representations from each self-attention layer and concatenate them pairwise to represent relations between them. Additionally, we leverage the last hidden representation in same manner. To effectively aggregate information, we apply a gated sum and then predict relation with a shallow relation head. 
    }
    \label{fig:framework}
    \vspace{-4mm}
\end{figure*}

\subsection{Preliminaries}

In this section, we first introduce the formulation of the SGG task and then provide a brief review of the one-stage object detector that serves as the basis for our study.

\subsubsection{Formulation}
Scene graph generation is a task of generating a scene graph $\mathcal{G} = \{\mathcal{V}, \mathcal{E}\}$ from an image, where $\mathcal{V}$ denotes the node set consisting of objects and $\mathcal{E}$ denotes the edge set that represents the relation between objects. Each object entity $v_i \in \mathcal{V}$ has an object category label $v_i^{c}$ from a set of object categories $\mathcal{
C}_v$ and box coordinates $v_i^{b}$. Each relation $e_j \in \mathcal{E}$ represents the $j$-th triplet ($s_j$, $p_j$, $o_j$), where subject $s_j$ and object $o_j$ indicate related object entities and predicate $p_j$ has a relation category label $p_j^{c}$ from a set of predicate categories $\mathcal{C}_p$. Generating $\mathcal{V}$ and $\mathcal{E}$ correspond to object detection and relation extraction, respectively. 

\subsubsection{One-stage Object Detector}

Our model framework is deeply influenced by DETR~\cite{carion2020end}, a one-stage object detector.
In DETR, representations for input images are learned through CNN and Transformer~\cite{vaswani2017attention} encoder. Transformer decoder enhances object queries by utilizing self- and cross-attention mechanisms.
The final object detection heads predict object category labels and box coordinates from the contextualized object queries.

\vspace{4pt}
\noindent
\textbf{Backbone.}
CNN backbone generates $C$-dimensional feature map $F \in \mathbb{R}^{C \times H_F \times W_F}$ from an input image $x_\text{img} \in \mathbb{R}^{3 \times H_0 \times W_0}$. For ResNet-50 backbone, $H_F$ and $W_F$ are set to $H_0/32$ and $W_0/32$, respectively. 

\vspace{4pt}
\noindent
\textbf{Transformer Encoder.}
Transformer encoder enhances image representations based on multi-head self-attention. After applying a $1 \times 1$ convolution to reduce dimension from $C$ to $d_\text{model}$, the feature map is flattened to get an input sequence with length $H_FW_F$ for the Transformer encoder. Additional positional encodings are used to reflect the spatial information of the feature maps.

\vspace{4pt}
\noindent
\textbf{Transformer Decoder.}
Transformer decoder takes $N$ object queries as input and returns their representations. Each object query detects an individual object, and $N$ is usually set large enough to cover all objects in the image.
Through alternating self-attention and cross-attention layers, object queries learn features of object candidates in the input image. 
Note that causal attention mask is not applied to the self-attention layer; therefore, fully connected $N \times N$ attention weights among the $N$ object queries are learned from the multi-head self-attention as follows:
\begin{equation}
    A_{h}^{l} = f(Q_{h}^{l}, K_{h}^{l}) = 
    \text{softmax}(Q_{h}^{l}{K_{h}^{l}}^T/\sqrt{d_{\text{head}}}),
\label{eq:self_attention}
\end{equation}
where $A_{h}^{l} \in \mathbb{R}^{N \times N}$ denotes the attention weights from the $h$-th head in the $l$-th layer. $Q_{h}^{l} \in \mathbb{R}^{N \times d_\text{head}}$ and $K_{h}^{l} \in \mathbb{R}^{N \times d_\text{head}}$ are attention queries and keys, respectively. 

\vspace{4pt}
\noindent
\textbf{Object Detection Heads.}
Object detection heads detect $N$ object candidates $\{\hat{v}_i\}_{i=1}^{N}$ from the last layer representations of the object queries $Z^L$. For each object query, a linear layer is used to predict the object category label $\hat{v}_i^c \in  \mathbb{R}^{N \times|\mathcal{C}_v|}$, and three-layer perceptron with ReLU activation is used to obtain the box coordinates $\hat{v}_i^b \in \mathbb{R}^{N \times4}$.

\vspace{4pt}
\noindent
\textbf{Object Detection Loss.}\label{object_detection}
To match $N$ detected object candidates $\{\hat{v}_i\}_{i=1}^{N}$ and $M$ ground truth objects $\{v_i\}_{i=1}^{M}$, Carion~\etal~\cite{carion2020end} pad the ground truth objects with $\phi$ (no object) and find the best permutation of predicted objects that minimizes total bipartite matching costs. From the permutated predictions $\{\hat{v}_i'\}_{i=1}^{N}$, the loss is calculated as follows:
\begin{equation}
    \mathcal{L}_\text{od} = \Sigma_{i=1}^N [\lambda_{c}\mathcal{L}_{c}(\hat{v}_i'^c, v_i^c) + \mathbbm{1}_{v_i^c\neq\phi}(\lambda_{b}\mathcal{L}_{b}(\hat{v}_i'^b, v_i^b))],
\label{eq:object_loss}
\end{equation}
where $\mathcal{L}_{c}$ and $\mathcal{L}_{b}$ denote loss function for category labels and box coordinates, respectively.

\subsection{EGTR}

We propose a novel lightweight relation extractor, EGTR, which exploits the self-attention of DETR decoder, as depicted in~\cref{fig:framework}.
Since the self-attention weights in~\cref{eq:self_attention} contain $N \times N$ bidirectional relationships among the $N$ object queries, our relation extractor aims to extract the predicate information from the self-attention weights in the entire $L$ layers, by considering the attention queries and keys as subjects and objects, respectively.

In order to preserve rich information learned in the self-attention layer, we devise another relation function $f$ between the attention queries and keys instead of the dot product attention in~\cref{eq:self_attention}.
As concatenation preserves representations completely, we concatenate the representations on all $N \times N$ pairs of attention queries and keys as shown in~\cref{fig:framework} (a) to get the relation representations of the $l$-th layer $R_a^l \in \mathbb{R}^{N \times N \times 2d_\text{model}}$. Before pairwise concatenation, to help queries and keys play the role of subjects and objects, a linear projection is added as follows:
\begin{equation}
    \begin{aligned}
        R_a^l &= [Q^l W_S^l; K^l W_O^l],\\
    \end{aligned}
\label{eq:relation_attention}
\end{equation}
where $Q^l \in \mathbb{R}^{N \times d_\text{model}}$ and $K^l \in \mathbb{R}^{N \times d_\text{model}}$ refer to attention queries and keys of the $l$-th layer, and $[\,\cdot\,;\cdot\,]$ denotes the pairwise concatenation. $W_S^l$ and $W_O^l$ are linear weights of shape $\mathbb{R}^{d_\text{model} \times d_\text{model}}$.

We also utilize the last layer representations of the object queries $Z^L \in \mathbb{R}^{N \times d_\text{model}}$, which are used for the object detection in the same manner:
\begin{equation}
    \begin{aligned}
        R_z = [Z^L W_S ; Z^L W_O],
    \end{aligned}
\label{eq:relation_final}
\end{equation}
where $W_S$, $W_O$ are linear weights of shape $\mathbb{R}^{d_\text{model} \times d_\text{model}}$.
To effectively use the various information learned in all layers, we introduce a gating mechanism as follows:
\begin{equation}
    \begin{aligned}
        g_a^l &= \sigma(R_a^l W_G), g_z = \sigma(R_z W_G),\\
    \end{aligned}
\label{eq:relation_gate}
\end{equation}
where $g_a^l$ and $g_z \in \mathbb{R}^{N \times N \times 1}$ represent the gate values obtained through a single linear layer $W_G \in \mathbb{R}^{2d_\text{model} \times1}$ for $R_a^l$ and $R_z$, respectively.
Finally, we extract the relation graph from the gated summation of the relation representations across all layers as follows:

\begin{equation}
    \begin{aligned}
        \hat{G} &= \sigma(\text{MLP}_\text{rel}(\Sigma_{l=1}^{L} (g_a^l * R_a^l) + g_z * R_z)), \\
    \end{aligned}
\label{eq:relation_graph}
\end{equation}
where $\hat{G} \in \mathbb{R}^{N\times N \times |\mathcal{C}_p|}$ denotes predicted relation graph and $\text{MLP}_\text{rel}$ is a three-layer perceptron with ReLU activation. Note that we use sigmoid function $\sigma$ so that multiple relationships can exist between objects.

\subsection{Learning and Inference}
To train EGTR, we perform multi-task learning. In addition to object detection and relation extraction, we devise connectivity prediction, an auxiliary task for relation extraction. The overall loss for the framework is as follows:
\begin{equation}
    \mathcal{L} = \mathcal{L}_\text{od} + \lambda_\text{rel}\mathcal{L}_\text{rel} + \lambda_\text{con}\mathcal{L}_\text{con},
\label{eq:loss}
\end{equation}
where $\mathcal{L}_\text{od}$ is object detection loss in~\cref{eq:object_loss}.
$\mathcal{L}_\text{rel}$ and $\mathcal{L}_\text{con}$ are loss functions for the relation extraction and connectivity prediction, respectively.
Since explicit object detection loss is used, EGTR has the capability to detect all nodes in the scene graph.
The details of the loss for each task are provided below.

\begin{figure}[t!]
     \centering
     \includegraphics[width=0.94\linewidth]{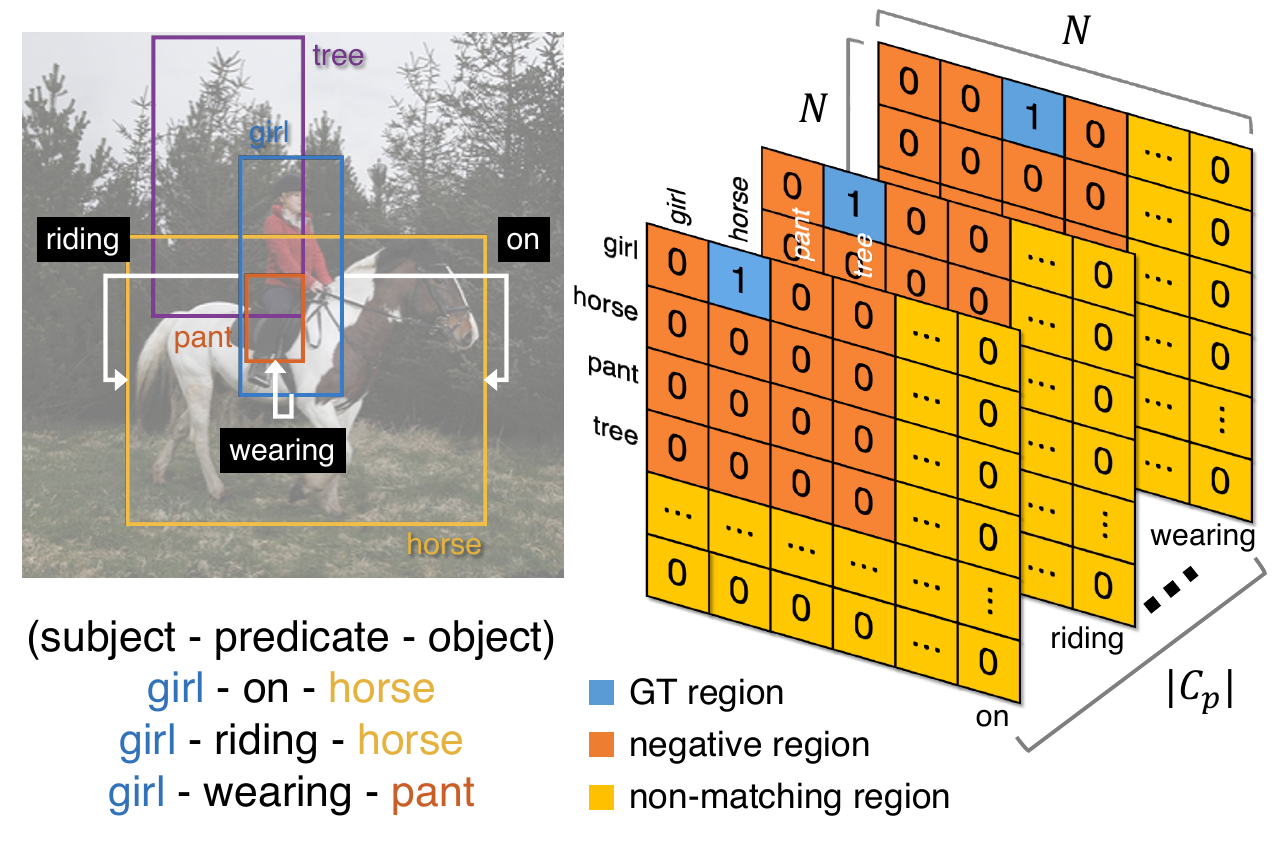}
     \caption{\textbf{Example of graph $G$ region.} The target graph has a shape of $N \times N \times |\mathcal{C}_p|$. The graph is severely sparse since the number of object queries $N$ is set large enough to cover objects in the image. We split $G$ into GT, negative, and non-matching regions.}
     \label{fig:graph}
     \vspace{-4mm}
\end{figure}

\subsubsection{Relation Extraction} 
We use binary cross-entropy loss for the relation extraction. To match predicted graph $\hat{G} \in \mathbb{R}^{N \times N \times |\mathcal{C}_p|}$ and ground truth triplet set $\mathcal{E}$, we first encode $\mathcal{E}$ as a one-hot ground truth graph $G \in \mathbb{R}^{N \times N \times |\mathcal{C}_p|}$ by padding the regions that do not correspond to relations between the ground truth objects as zero. Then, we permutate indices of the predicted graph using the permutation found in the object detection. From the permutated graph $\hat{G}'$, the loss for the relation extraction is calculated as $\mathcal{L}_\text{rel} = \mathcal{L}_{r}(\hat{G}', G)$.

However, since the size of the graph is proportional to the square of $N$, the sparsity of the ground truth graph $G$ is too severe. For instance, the density is only $10^{-14}$ when $N$ is set to $200$ for the Visual Genome~\cite{krishna2017visual} validation dataset. Therefore, we divide $G$ into three regions: (1) GT region, (2) negative region, and (3) non-matching region, as shown in~\cref{fig:graph}.
The GT region indicates the ground truth triplets where the value of $G$ is $1$.
The negative region consists of the triplets in which both subjects and objects are composed of the ground truth objects, but no relation exists between them.
The non-matching region represents the zero-padded region. It is paired with a region consisting of the object candidates that do not match the ground truth objects and match $\phi$ in~\cref{eq:object_loss}.
For each region, we apply different techniques to effectively train the relation extraction with object detection as subsequently described.

\vspace{4pt}
\noindent
\textbf{Adaptive Smoothing.}
We propose a novel adaptive smoothing for the GT region. 
For $G_{ijk}$ belonging to the GT region, the model is trained to predict the $k$-th predicate category between subject entity $v_i$ and object entity $v_j$. However, since $\hat{v}_i'$ and $\hat{v}_j'$, which match $v_i$ and $v_j$ respectively, do not have enough representations about the corresponding ground truth object at the beginning of the training, it may not be appropriate to predict the probability of the predicate as $1$. 
Moreover, even when the object detection performance is reasonably assured, the detection performance may still vary for individual object candidates.
Therefore, we reflect the detection performance of each object candidate on the relation label via adaptive smoothing.

We first measure the uncertainty of each object candidate with the corresponding bipartite matching cost. For object candidate $\hat{v}_i'$, we define the uncertainty as follows:
\begin{equation}
    u_i = \sigma(\text{cost}_i - \text{cost}_\text{min} + \sigma^{-1}(\alpha)),
\label{eq:uncertainty}
\end{equation}
where $\text{cost}_i$ denotes the matching cost and $\text{cost}_{\text{min}}$ indicates the matching cost when $\hat{v}_i'$ perfectly matches $v_i$. $\alpha$ is a nonnegative hyperparameter representing the minimum uncertainty. We set the value of $G_{ijk}$ to $(1-u_i)(1-u_j)$ taking the uncertainty into account. 
By using the relation label adjusted by the uncertainty, the multi-task learning of object detection and relation extraction is dynamically regulated according to the quality of the detected objects.

\vspace{4pt}
\noindent
\textbf{Negative and Non-matching Sampling.}
Rather than employing all negatives, we sample them from the negative region. Inspired by the hard negative mining introduced by Liu~\etal~\cite{liu2016ssd}, we sort all negatives based on the predicted relation score $\hat{G}'_{ijk}$ and choose the top $k_\text{neg} \times |\mathcal{E}|$ most challenging negatives. Similarly, we extract $k_\text{non} \times |\mathcal{E}|$ hard samples from the non-matching region. As the non-matching region typically encompasses a substantial part of the graph $G$, this approach notably diminishes sparsity.

\subsubsection{Connectivity Prediction}

Influenced by Graph-RCNN~\cite{yang2018graph} that predicted a relatedness to prune object pairs, we propose a connectivity prediction that predicts whether at least one edge between two object nodes exists for the relation extraction. 
We get a connectivity graph $\hat{E} \in \mathbb{R}^{N \times N \times 1}$ in a similar way to get the relation graph in~\cref{eq:relation_graph} with the same relation source representations. Instead of $\text{MLP}_{\text{rel}}$ for multi-label prediction, we use another $\text{MLP}_\text{con}$ for binary classification.
We calculate the binary cross entropy loss from the permutated connectivity graph as follows: $\mathcal{L}_\text{con} = \mathcal{L}_{r}(\hat{E}', E)$. 

\subsubsection{Inference}

For model inference, we get triplet scores by multiplying predicate score $\hat{G}_{ijk}$ by the corresponding class scores of $\hat{v}_i^c$ and $\hat{v}_j^c$. Note that we set $\hat{G}_{iik}$ to $0$ to prevent self-connections in which the subject and object are the same entity.
In addition, we enhance the triplet scores by utilizing the connectivity score $\hat{E}_{ijk}$. 
By multiplying the connectivity score, we can effectively filter out the triplets that are not likely to have a relation between the subject and object.

%% file: tex/4_experiment.tex
\input{table/result_vg}

\subsection{Datasets and Evaluation Settings}
We conduct experiments on two SGG datasets. We describe the datasets and evaluation settings for each dataset. Detailed settings for each dataset are presented in the supplementary materials.

\vspace{4pt}
\noindent
\textbf{Visual Genome.} 
Visual Genome~\cite{krishna2017visual} is the most representative SGG dataset, consisting of $57\text{K}$ training images, $5\text{K}$ validation images, and $26\text{K}$ test images. We follow the popular Visual Genome split, retaining the most frequent $150$ object categories and $50$ relation categories. We adopt Scene Graph Detection (SGDet) evaluation settings and report Recall@$k$ (R@$k$) which is class agnostic and mean Recall@$k$ (mR@$k$) that aggregates the recalls for each predicate category. 
Following Motifs~\cite{zellers2018neural}, these metrics are measures with graph constraint, which means each object pair can have a single predicate category. 
Since they are only related to triplet detection of the scene graph, we report $\text{AP}50$ that evaluates the detection performance of all objects appearing in the scene.
Furthermore, we report the number of model parameters and Frames Per Second (FPS) to measure efficiency.

\vspace{4pt}
\noindent
\textbf{Open Image V6.} 
Open Image V6 ~\cite{kuznetsova2020open} is also widely used dataset, comprising of $126\text{K}$ images for training, $2\text{K}$ for validation, and $5\text{K}$ for test set. It includes $601$ object categories and $30$ relation categories. 
For evaluation, we adopt both recall and weighted mean AP (wmAP) following standard settings. For recall evaluation, micro-R@$50$ is adopted.
The wmAP is evaluated with two settings: wmAP$_\text{rel}$ for predicting boxes of subject entity and object entity separately and wmAP$_\text{phr}$ for predicting a union box of them.
The final score is calculated by $0.2 \times \text{micro-R@}50 + 0.4 \times \text{wmAP}_\text{rel} + 0.4 \times \text{wmAP}_\text{phr}$.

\input{table/result_oi}

\subsection{Implementation Details}

We employ Deformable DETR~\cite{zhu2020deformable} that improves the convergence speed of the DETR with ResNet-50~\cite{he2016deep} as a backbone. It is worth noting that our approach can be extended to any object detector that incorporates self-attention mechanisms between object features, including models like DETR~\cite{carion2020end}, Sparse R-CNN~\cite{sun2021sparse}, and others. 
We follow the configurations of the original Deformable DETR, except that we use only $200$ object queries.
To speed up convergence, we first train the object detector with the target dataset and subsequently train the SGG task using the pre-trained object detector, similar to prior work~\cite{li2022sgtr}. 
To calculate the overall loss in Equation~\ref{eq:loss}, $\lambda_\text{rel}$ is set to $15$. $\lambda_\text{con}$ is set to $30$ and $90$ for the Visual Genome and Open Image V6, respectively. We set $\alpha$ as $10^{-14}$ for adaptive smoothing and both $k_\text{neg}$ and $k_\text{non}$ are set to $80$.

\subsection{Results}

We present quantitative results and perform a comparative analysis of our proposed framework with representative SGG models.
Additional experimental results are included in the supplementary materials.

\vspace{4pt}
\noindent
\textbf{Visual Genome.}
Visual Genome results are shown in~\cref{table:vgtable}.
Our proposed method demonstrates competitive performance with the current one-stage SGG models that have $1.5$ to $6.5$ times larger parameters with the fastest inference speed.
In particular, our method achieves the highest object detection performance and shows comparable performance in the triplet detection compared to SSR-CNN~\cite{teng2022structured}, the state-of-the-art method. 
The results demonstrate that our method can generate scene graphs in an efficient and effective way by exploiting the by-products of the object detector.
By applying logit adjustment~\cite{teng2022structured}, a technique for predicting tail predicate classes well, our method performs favorably in the trade-off between R@$k$ and mR@$k$, showing superiority over existing state-of-the-art models.
Note that for triplet detection models~\cite{khandelwal2022iterative, teng2022structured} without an explicit object detector, AP$50$ is measured by applying non-maximum-suppression (NMS) to the union of predicted subjects and objects set. AP$50$ performance is discussed further in the section~\ref{sec:discussion}.

\vspace{4pt}
\noindent
\textbf{Open Image V6.}
As depicted in~\cref{table:oitable}, the experiments conducted on the Open Image V6 dataset also demonstrate competitive performance, underscoring the effectiveness and robustness of our method across different datasets.

\subsection{Ablation Studies}
We analyze the effects of our model components. All of the ablation studies are done with the Visual Genome dataset.

\vspace{4pt}
\noindent
\textbf{Relation Sources.}
In~\cref{table:relpredictortable}, we investigate the source of relations used for the relation extractor.
To verify the benefits of using attention queries and keys, we change $Q^l$ and $K^l$ in~\cref{eq:relation_attention} to $Z^l$, the hidden states of each layer. Remarkably, when simply using the hidden states of all layers, the triplet detection performance decreases. Surprisingly, the performance is lower than that of using only $R_z$ from the last layer. 
On the other hand, when only attention by-products are used as a relation source, the performance is similar to that of using only hidden states of the last layer, the most contextualized representations for detecting objects.
Incorporating both attention by-products and final hidden states yields improved performance by amalgamating varied information. The results underscore that the attention by-products of the object detectors contain rich information for relation extraction.

\vspace{4pt}
\noindent
\textbf{Training Techniques.}
In~\cref{table:mainablation}, we ablate proposed techniques. Without any techniques, our model framework shows low performance. When each technique is applied, it substantially improves both R@$50$ and mR@$50$, and the best performance is achieved when all techniques are combined. The results demonstrate that our proposed techniques help to train the model framework effectively. 
Additional experimental results are provided in the supplementary materials.

\input{table/ablation_f}

\input{table/ablation_smoothing}

\input{table/ablation_sampling}

\vspace{4pt}
\noindent
\textbf{Sampling Methodology.}
In~\cref{table:sortablation}, we investigate the impact of hard sampling methods for the negative and non-matching regions.
Hard negatives and hard non-matchings show a trade-off that improves the performance of R@$50$ and mR@$50$, respectively, and additional performance improvements are achieved for both measures when they are used together. 
For hard negatives, we speculate that R@$50$ performance is improved by selecting negatives for the tail predicate class, which is likely to exist in reality but is not in the annotation. 
For hard non-matchings, mR@$50$ performance seems to have improved by choosing object candidates that are akin to the ground truth objects but unmatched with them.
This encourages the model to predict diverse predicate classes by preventing duplicate object pairs, which are likely to predict head predicate classes, from being predicted.

\subsection{Discussion}\label{sec:discussion}

Regarding the highest AP50 score, we posit that it is because our proposed method focuses on not only objects with relations but also objects without any relations. To validate this, we calculate AP$50$ for two subsets of the ground-truth objects: those with relations and those without relations. 
We compare our method with two triplet detection-based models.~\cref{table:ap} illustrates that our method achieves significantly high $\text{AP}50_\text{no-rel}$.
Considering that triplet detection performance is related to $\text{AP}50_\text{rel}$, while triplet detection-based models only focus on objects with relations to detect triplets well, our method is capable of extracting the complete scene graph, including objects without relations.

\input{table/ap50_ablation}

\begin{figure}[t]
  \centering
  \begin{subfigure}{0.234\textwidth}
    \centering
    \includegraphics[width=\linewidth]{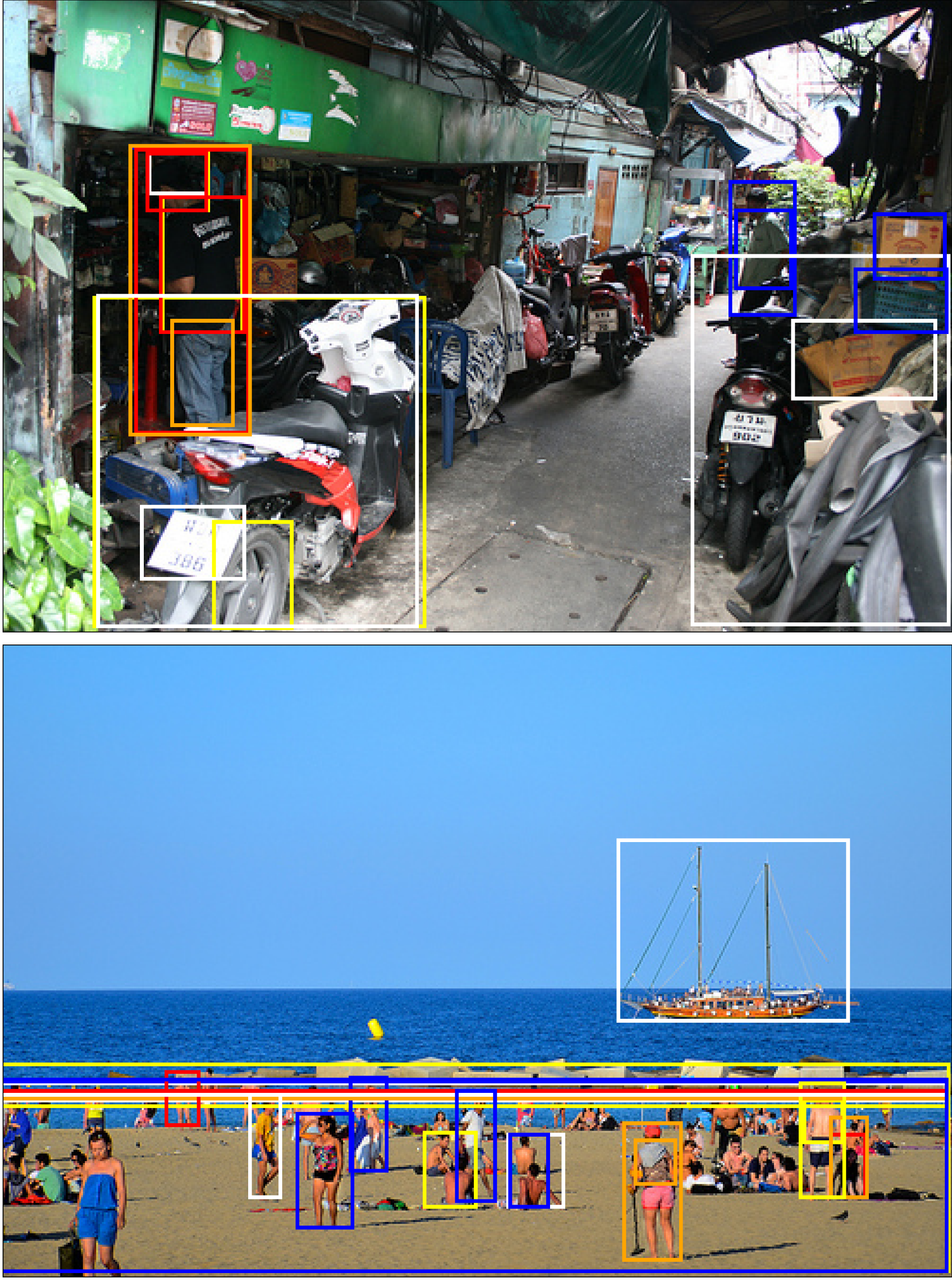}
    \caption{SSR-CNN~\cite{teng2022structured}}
    \label{fig:subfig1}
  \end{subfigure}
  \begin{subfigure}{0.234\textwidth}
    \centering
    \includegraphics[width=\linewidth]{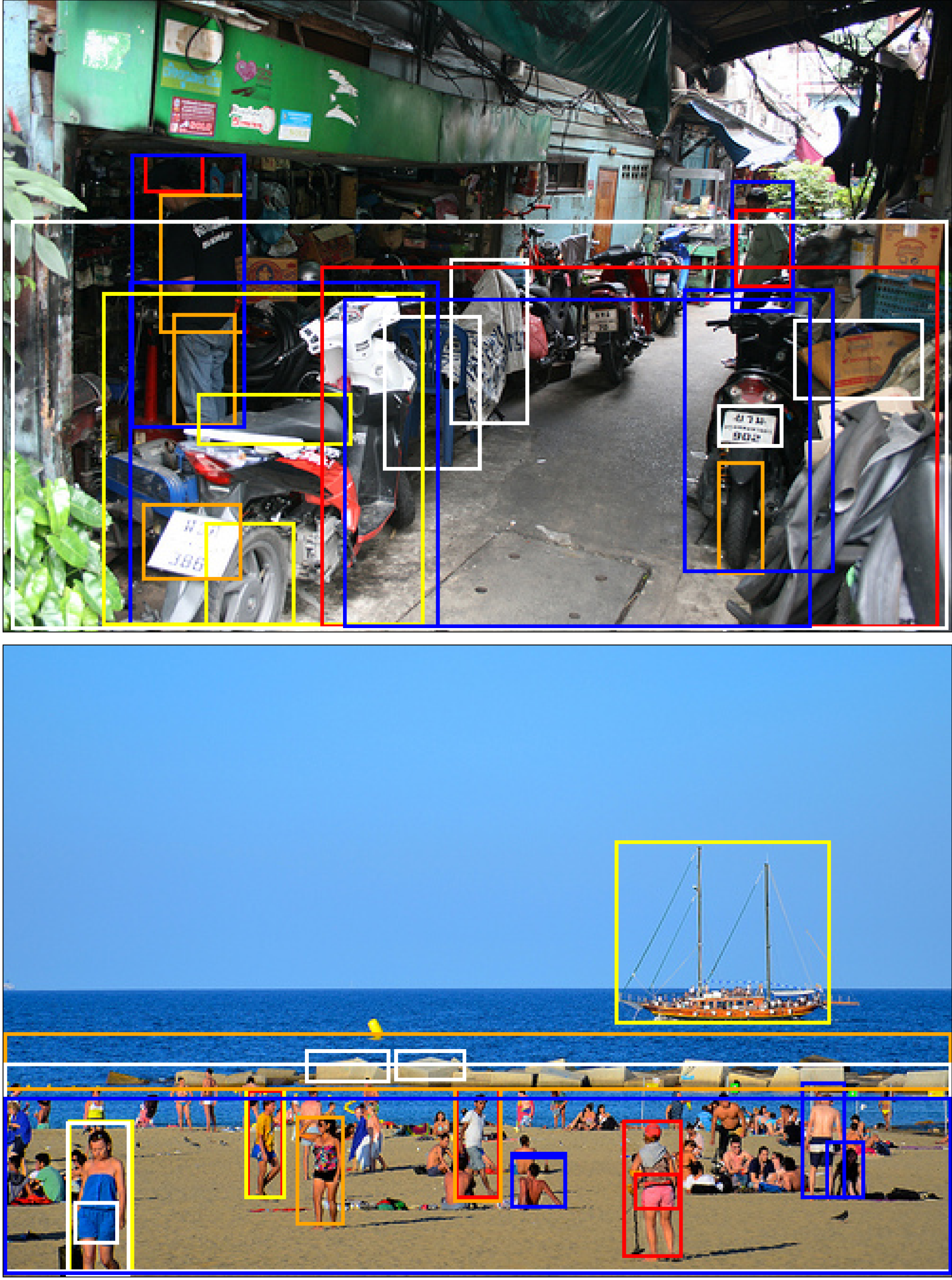}
    \caption{EGTR (Ours)}
    \label{fig:subfig2}
  \end{subfigure}
  \vspace{-.5em}
  \caption{\textbf{Comparison of detected subjects and objects.} We select the top $20$ predictions of SSR-CNN and EGTR and visualize bounding boxes of detected subjects and objects.}
  \label{fig:bboxprediction}
  \vspace{-4mm}
\end{figure}

For further analysis, we visualize bounding boxes for subjects and objects from the top $20$ predictions in~\cref{fig:bboxprediction}. 
For SSR-CNN~\cite{teng2022structured}, which directly predicts triplets, we observe a significant number of overlapping bounding boxes. 
On the other hand, our proposed model leverages predictions from an explicit object detector, reducing redundant predictions. Furthermore, our method detects various subjects and objects that appear in the scene rather than focusing on objects that are more likely to have relations.

%% file: table/result_vg.tex
\begin{table*}[t!]
\centering
\footnotesize
    \begin{center}
        \begin{tabular*}{\linewidth}{@{\extracolsep{\fill}} l|l|cc|c|ccc|ccc }
            \toprule
            &Model&  \# params (M) & FPS  & AP$50$ & R@$20$ & R@$50$ & R@$100$ & mR@$20$ & mR@$50$ & mR@$100$    \\
            \midrule
            \multirow{8}{*}{\rotatebox[origin=c]{90}{two-stage}}  
            &IMP (EBM)~\cite{xu2017scene, suhail2021energy}& \textit{322.2} & \textit{2.0} & 28.1 &  18.1 & 25.9 & 31.2 & 2.8 & 4.2 & 5.4  \\
            &VTransE~\cite{zhang2017visual} & \textit{312.3} & \textit{3.5} & - & 24.5 & 31.3 & 35.5 & 5.1 & 6.8 & 8.0 \\
            & Motifs~\cite{zellers2018neural} & \textit{369.9} & \textit{1.9} & 28.1  & 25.1 & 32.1 & 36.9 & 4.1 & 5.5 & 6.8 \\
            &VCTree~\cite{tang2019learning} &\textit{361.5}& \textit{0.8} & 28.1 & 24.8 & 31.8 & 36.1 & 4.9 & 6.6 & 7.7\\
            &VCTree (TDE)~\cite{tang2019learning, tang2020unbiased} & \textit{361.3} & \textit{0.8} & 28.1 & 14.0 & 19.4 & 23.2 & 6.9 & 9.3 & 11.1\\
            &VCTree (EBM)~\cite{tang2019learning, suhail2021energy} & \textit{372.5} & - & 28.1 & 24.2 & 31.4 & 35.9 & 5.7 & 7.7 & 9.1 \\ 
            &GPS-Net~\cite{lin2020gps} & - & -  &-& - & 31.1 & 35.9 & - & 6.7 & 8.6\\
            &BGNN~\cite{li2021bipartite} & 341.9 & \textit{1.7}&29.0& 23.3 & 31.0 & 35.8 & 7.5 & 10.7 & 12.6\\
            \midrule
            \multirow{11}{*}{\rotatebox[origin=c]{90}{one-stage}}   &FCSGG~\cite{liu2021fully} & 87.1 & \textit{6.0}& \underline{28.5} &16.1 & 21.3 & 25.1 & 2.7 & 3.6 & 4.2 \\
            &RelTR~\cite{hu2018relation}& \underline{63.7} & \textit{\underline{13.4}}  & 26.4 & 21.2 & 27.5 & - & 6.8 & 10.8 & - \\
            &SGTR~\cite{li2022sgtr} & \textit{117.1} & \textit{6.2} & 25.4 & - & 24.6 & 28.4 & - & 12.0 & 15.2 \\
            &Relationformer~\cite{shit2022relationformer}  & 92.9 & \textit{8.5}& 26.3& 22.2 & 28.4 & 31.3 & 4.6 & 9.3 & 10.7 \\
            
            &Iterative SGG~\cite{khandelwal2022iterative} & \textit{93.5} & \textit{6.0} &\textit{27.7}\dag & - &29.7 & 32.1 & - & 8.0 & 8.8   \\
            &SSR-CNN~\cite{teng2022structured} &\textit{274.3} & \textit{4.0} & \textit{23.8}\dag & \textbf{25.8} & \textbf{32.7} & \textbf{36.9} & 6.1 & 8.4 & 10.0 \\
            &SSR-CNN~\cite{teng2022structured} $_{\text{LA}, \tau =0.3}$  & \textit{274.3}& \textit{4.0}& \textit{23.8}\dag & 18.4 & 23.3 & 26.5 & \textbf{13.5} & \textbf{17.9} & \underline{21.4}  \\
            \cmidrule{2-11}
            &\textbf{EGTR} (Ours)& \textbf{42.5} &  \textbf{14.7}& \textbf{30.8}& \underline{23.5}  & \underline{30.2} & \underline{34.3} & 5.5 & 7.9 & 10.1  \\ 
            &\textbf{EGTR} (Ours) $_{\text{LA},\tau =0.7}$& \textbf{42.5} &  \textbf{14.7} & \textbf{30.8}& 15.7 & 18.7 & 20.5 & \underline{12.1} & \underline{17.8} & \textbf{21.7}  \\ 
            &\textbf{EGTR} (Ours) $_{\text{LA}, \tau =0.5}$& \textbf{42.5} &  \textbf{14.7} & \textbf{30.8}& 19.7 & 24.2 & 26.7 & 11.0 & 17.1 & \underline{21.4}  \\ 
            &\textbf{EGTR} (Ours) $_{\text{LA}, \tau =0.3}$& \textbf{42.5} &  \textbf{14.7} & \textbf{30.8}& 22.4 & 28.2 & 31.7 & 8.8 & 14.0 & 18.3  \\ 
            \bottomrule
        \end{tabular*}
    \end{center}
    \vspace{-1.5em}
    \caption{\textbf{Graph-Constraint results on the test set of Visual Genome dataset.} We report the results of representative two-stage SGG models based on the Faster R-CNN~\cite{ren2015faster} object detector with a ResNeXt-101-FPN~\cite{lin2017feature,xie2017aggregated} backbone and concurrent one-stage SGG models. 
    Among the one-stage models, the best is highlighted in bold, and the second-best is indicated with underlining. 
    $\text{LA}$ denotes logit adjustment proposed in SSR-CNN~\cite{teng2022structured}. \dag~represents that we concatenate the predicted subjects and objects and then apply non-maximum-suppression with a threshold of $0.3$. \textit{Italic} indicates the evaluation of metrics using publicly available model checkpoints. FPS is measured with a single V$100$ for images resized to a minimum of $600$ for the shortest side and a maximum of $1000$ for the longest side.
    }
    \label{table:vgtable}
    \vspace{-4mm}
\end{table*}

%% file: table/result_oi.tex
\begin{table}[t!]
\footnotesize
\centering
    \setlength{\tabcolsep}{0.5\tabcolsep}
    \begin{center}
        \begin{tabular*}{\linewidth}{@{\extracolsep{\fill}} lcccc }
            \toprule
            Model & score &  micro-R@$50$ & $\text{wmAP}_\text{rel}$ & $\text{wmAP}_\text{phr}$ \\ 
            \midrule
            Motifs~\cite{zellers2018neural} & 38.9 & 71.6 & 29.9 & 31.6 \\
            VCTree~\cite{tang2019learning} & 40.2 & 74.1 & 34.2 & 33.1 \\
            GPS-Net~\cite{lin2020gps}& 41.7 & 74.8 & 32.9 & 34.0 \\
            BGNN~\cite{li2021bipartite} & 42.1 & 75.0 & 33.5 & 34.2 \\
            \midrule
            RelTR~\cite{hu2018relation} & 43.0 & 71.7 & 34.2 & 37.5 \\
            SGTR~\cite{li2022sgtr} & 42.3 & 59.9 & 37.0 & 38.7 \\
            SSR-CNN~\cite{teng2022structured}  & \textbf{49.4} & \textbf{76.7} & \underline{41.5} & \textbf{43.6} \\
            \midrule
            \textbf{EGTR} (Ours) & \underline{48.6} & \underline{75.0} & \textbf{42.0} & \underline{41.9} \\
            \bottomrule
        \end{tabular*}
    \end{center}
    \vspace{-1.5em}
    \caption{\textbf{Results on test set of Open Image V6.} 
    The score is a weighted sum of $\text{micro-R@}50$, $\text{wmAP}_\text{rel}$, and $\text{wmAP}_\text{phr}$. 
    }
    \vspace{-4.mm}
    \label{table:oitable}
\end{table}

%% file: table/ablation_f.tex
\begin{table}[t!]
\footnotesize
\centering
    \begin{center}
        \setlength{\tabcolsep}{0.8\tabcolsep}
        \begin{tabular*}{\linewidth}{@{\extracolsep{\fill}}ccccc}
            \toprule
            $R_a^l$ source &$R_a^l$& $R_z$ &R@$50$ &mR@$50$ \\
            \midrule
            $Q^l$ \& $K^l$ & \textbf{\checkmark} & \textbf{\checkmark} & \textbf{30.2} & \textbf{7.9} \\
             $Z^l$ & \checkmark& \checkmark & 29.6 & 7.4 \\
             - & & \checkmark &   29.9 & 7.6 \\
             $Q^l$ \& $K^l$ &\checkmark &  &   29.8 & 7.7 \\
            
            \bottomrule
        \end{tabular*}
    \end{center}
    \vspace{-1.5em}
    \caption{\textbf{Ablation study on relation source.}}
    \label{table:relpredictortable}
\end{table}

%% file: table/ablation_smoothing.tex
\begin{table}[t!]
\footnotesize

\centering
    \begin{center}
        \begin{tabular*}{\linewidth}{@{\extracolsep{\fill}} ccccc }
            \toprule
            adaptive smoothing & $\mathcal{L}_\text{con}$ & sampling & R@$50$ & mR@$50$  \\
            \midrule
             & & & 26.6 & 5.3       \\        
            \checkmark & & & 28.3 & 6.5 \\
             & \checkmark & & 29.6 & 7.0 \\
             & & \checkmark & 28.9 & 7.1 \\
            \checkmark & \checkmark & \checkmark &\textbf{30.2} & \textbf{7.9} \\
            \bottomrule
        \end{tabular*}
    \end{center}
    \vspace{-1.5em}
    \caption{\textbf{Ablation study on proposed techniques.} }
    \label{table:mainablation}
\end{table}

%% file: table/ablation_sampling.tex
\begin{table}[t!]
\footnotesize
\centering
    \begin{center}
        \begin{tabular*}{\linewidth}{@{\extracolsep{\fill}} cccc }
            \toprule
              hard negative & hard non-matching  & R@$50$ & mR@$50$  \\
            \midrule
             \textbf{\checkmark}& \textbf{\checkmark}& \textbf{30.2 }&\textbf{7.9} \\

             \checkmark && 30.0 & 7.1 \\
            & \checkmark & 29.6 & 7.7 \\
            \bottomrule
        \end{tabular*}
    \end{center}
    \vspace{-1.5em}
    \caption{\textbf{Ablation study on sampling methods.} }
    \label{table:sortablation}
    \vspace{-4mm}
\end{table}

%% file: table/ap50_ablation.tex
\begin{table}[t!]
\footnotesize
\centering
    \begin{center}
        \begin{tabular*}{\linewidth}{@{\extracolsep{\fill}} lccc }
            \toprule
            Model & $\text{AP}50$ & $\text{AP}50_\text{rel}$  & $\text{AP}50_\text{no-rel}$ \\
            \midrule
            Iterative SGG~\cite{khandelwal2022iterative}\dag & 27.7 &\textbf{24.3} & 7.8  \\
            SSR-CNN~\cite{teng2022structured}\dag & 23.8 & 20.2 & 7.4\\
            \textbf{EGTR} (Ours) & \textbf{30.8} & \textbf{24.3} & \textbf{10.7}\\  
            \bottomrule
        \end{tabular*}
    \end{center}
    \vspace{-1.5em}
    \caption{\textbf{AP50 for two subsets of objects.} $\text{AP}50_\text{rel}$ and $\text{AP}50_\text{no-rel}$ denote scores evaluated only on objects having at least one edge and no edge, respectively. Note that each measure operates on its unique scale since the ground-truth objects used for each measure differ. \dag~indicates that additional NMS is applied to the union of subjects and objects set.
    }
    \label{table:ap}
\end{table}

%% file: tex/5_conclusion.tex
In this study, we propose the lightweight one-stage scene graph generator EGTR. We significantly reduce the model complexity by harnessing the relationships among the object queries learned from the self-attention layers of the object decoder. 
Furthermore, we devise a novel adaptive smoothing technique that helps multi-task learning of object detection and relation extraction by adjusting the relation label according to the object detection performance. As an auxiliary task of relation extraction, connectivity prediction contributes to the effective learning of EGTR. 
We conduct extensive experiments and ablation studies, and the results demonstrate that EGTR achieves the highest object detection performance and competitive triplet detection capabilities with the fastest inference speed.

\vspace{4pt}
\noindent
\textbf{Acknowledgements.} We thank Taeho Kil, Geewook Kim, and the anonymous reviewers for their insightful comments and suggestions. 

%% file: tex/7_contribution.tex
\textbf{Jinbae Im} initiated and led the project, proposed the main ideas, and made significant contributions throughout the process, including implementation, experiments, and manuscript writing.
\textbf{JeongYeon Nam} implemented experimental ideas, conducted a major part of the experiments, and made significant contributions to the manuscript writing and the development of the paper's direction.
\textbf{Nokyung Park} managed the reproduction and validation of other models, conducted performance evaluations and comparisons, and contributed to enhancing the model's performance and manuscript writing.
\textbf{Hyungmin Lee} designed and conducted a proof of concept experiment associated with the model architecture and implemented and conducted experiments mainly related to relation prediction.
\textbf{Seunghyun Park} co-initiated the project, advised it from its inception, participated in manuscript writing, and significantly contributed to shaping the project's direction as a senior researcher.

%% file: tex/6_supplementary.tex
\section*{Overview of Supplementary Material}
This supplementary material provides detailed information not covered in the main manuscript due to space constraints. First,~\cref{sup:method} describes the bipartite matching and the detailed loss function for object detection. Then, in~\cref{sup:eval}, we introduce the evaluation methods for each dataset used in our study: Visual Genome and Open Image V6.~\cref{sup:implementation} provides the implementation details. Finally,~\cref{sup:results} shows the results of various additional experiments.

\section{Method Details}\label{sup:method}

\subsection{Bipartite Matching}

We apply the bipartite matching used in DETR~\cite{carion2020end} to match $N$ predicted objects set $\{\hat{v}_i\}_{i=1}^{N}$ and $M$ ground truth objects set $\{v_i\}_{i=1}^{M}$.
Since $N$ is set large enough to handle all objects appearing in the image, we pad the ground truth objects with $\phi$ (no object). Subsequently, we find the best permutation $\sigma$ of $N$ predicted objects that minimizes the bipartite matching costs as follows:
\begin{equation}
    \sigma = \operatorname*{argmin}_{\hat{\sigma} \in \mathcal{S}_N} \sum_{i=1}^N \mathcal{L}_\text{match}(v_i, \hat{v}_{\hat{\sigma}(i)}),
\label{eq:bipartite}
\end{equation}
where $\mathcal{S}_N$ denotes all possible permutations of $N$ predicted objects. From the best permutation $\sigma$, we denote the permutated predictions as $\{\hat{v}_i'\}_{i=1}^{N}$, where $\hat{v}_i' = \hat{v}_{\sigma(i)}$.
The matching cost $\mathcal{L}_\text{match}$ is defined as $\mathcal{L}_\text{match}(v_i, \hat{v}_i') = \lambda_\text{c}\mathcal{L}_\text{match}^c(v_i^c, \hat{v}_i'^c)  + \lambda_\text{b}\mathcal{L}_\text{match}^b(v_i^b, \hat{v}_i'^b)$ with class matching cost $\mathcal{L}_\text{match}^c$ and box matching cost $\mathcal{L}_\text{match}^b$. Note that the matching cost is not considered when $v_i^c$ is $\phi$. $\mathcal{L}_\text{match}^c$ is defined as negative likelihood and $\mathcal{L}_\text{match}^b$ is defined as $\mathcal{L}_\text{match}^b(v_i^b, \hat{v}_i'^b) = \lambda_\text{IoU}\mathcal{L}_\text{IoU}(v_i^b, \hat{v}_i'^b) +  \lambda_\text{L1}{||v_i^b, \hat{v}_i'^b||}_1$, where $\mathcal{L}_\text{IoU}$ indicates generalized intersection over union (IoU) loss~\cite{rezatofighi2019generalized}. Note that $\mathcal{L}_\text{match}^c$ is changed to fit the focal loss~\cite{lin2017focal} for Deformable DETR~\cite{zhu2020deformable}. 

\subsection{Object Detection Loss}

We use the bipartite matching loss proposed in DETR for object detection. From the permutated predictions $\{\hat{v}_i'\}_{i=1}^{N}$, we compute the loss for $N$ matched pairs of the predicted objects and the ground truth objects as follows:
\begin{equation}
    \mathcal{L}_\text{od} = \sum_{i=1}^N [\lambda_{c}\mathcal{L}_{c}(v_i^c, \hat{v}_i'^c) + \mathbbm{1}_{v_i^c\neq\phi}(\lambda_{b}\mathcal{L}_{b}(v_i^b, \hat{v}_i'^b))],
\label{eq:object_loss_sup}
\end{equation}
where the loss consists of class loss $\mathcal{L}_c$ and box loss $\mathcal{L}_b$ that is the same as box matching cost $\mathcal{L}_\text{match}^b$. For $\mathcal{L}_c$, cross-entropy loss is used for DETR, and focal loss is used for Deformable DETR.

\section{Evaluation Details}\label{sup:eval}

\subsection{Visual Genome}

Among various evaluation methods for the scene graph generation task, including scene graph detection (SGDET), scene graph classification (SGCLS), and predicate classification (PRDCLS)~\cite{xu2017scene}, we adopt the SGDET evaluation protocol due to its rigorous and comprehensive nature compared to other methods. Unlike SGCLS or PRDCLS, where ground truth categories or box coordinates of objects are given, SGDET evaluates the performance of entity categories and box coordinates for each subject and object, as well as predicate categories collectively. We use the widely adopted value of $50\%$ as the thresholding parameter for IoU incorporated in SGDET. We also adopt the graph constraint evaluation method proposed in Zellers~\etal~\cite{zellers2018neural}, which enforces a limit of one predicted predicate between a given subject and object entity. We select the top 1 predicate for each object pair, as determined by multiplying the predicate score $\hat{G}_{ijk}$ by the connectivity score $\hat{E}_{ij}$ and multiplying the corresponding class scores of the subject $\hat{v}_i^c$ and object entity $\hat{v}_j^c$. We use recall at $k$ (R@$k$) and mean recall at $k$ (mR@$k$)~\cite{tang2019learning} as evaluation metrics. mR@$k$ is a balanced version of R@$k$ in that mR@$k$ can compensate for the bias of predicate categories by aggregating for all predicate categories.

\subsection{Open Image V6}

Open Image V6 is also assessed in the SGDET setting. We adopt recall and weighted mean AP (wmAP) following the standard settings proposed in Zhang~\etal~\cite{zhang2019graphical}. For recall, micro-R@50 is used following previous studies~\cite{li2021bipartite,cong2023reltr, li2022sgtr, teng2022structured}. For wmAP considering the ratio of each predicate category as the weight, wmAP of relationships ($\text{wmAP}_\text{rel}$) and wmAP of phrases ($\text{wmAP}_\text{phr}$) are adopted. $\text{wmAP}_\text{rel}$ evaluates whether both the subject entity and object entity boxes have IoU greater than $50\%$ with the corresponding ground truth boxes. $\text{wmAP}_\text{rel}$ evaluates the single box that encloses the boxes of the subject entity and the object entity. The final score is calculated by $0.2 \times \text{micro-R@}50 + 0.4 \times \text{wmAP}_\text{rel} + 0.4 \times \text{wmAP}_\text{phr}$. During model inference, we select the top $2$ predicates for each object pair following previous works~\cite{li2022sgtr,teng2022structured}.

\section{Implementation Details}
\label{sup:implementation}

Our object detector backbone is based on Deformable DETR~\cite{zhu2020deformable} with ResNet-$50$~\cite{he2016deep}.
To shorten the convergence time, we pre-train the object detector backbone using $8$ V$100$ GPUs with a batch size of $32$, employing AdamW~\cite{loshchilov2017decoupled} optimizer with a default learning rate of $10^{-4}$ and a decreased learning rate of $10^{-5}$ for ResNet-$50$.
EGTR is trained on $8$ V$100$ GPUs with a batch size of $64$, using a learning rate of $2\times10^{-4}$ for the relation extractor and scaling down the learning rates for the object detector and ResNet-$50$ by $100$ and $1000$ times, respectively.
Following the original DETR training scheme, we adopt a learning rate schedule that reduces the learning rate by a factor of $10$ after the model has trained to some extent.
Instead of a fixed learning schedule, we apply an adaptive schedule through early stopping.
For the object detector pretraining, we set the maximum number of epochs to $150$ for the first schedule and $50$ for the second schedule.
In the main EGTR training, we configure the first schedule for $50$ epochs and the second schedule for $25$ epochs.
For the hyperparameters of the bipartite matching and object detection loss, we follow the configurations of the original Deformable DETR.
$\lambda_\text{c}$, $\lambda_\text{b}$, $\lambda_\text{IoU}$, and $\lambda_\text{L1}$ are set to $2$, $1$, $2$, and $5$, respectively.
For Open Images V$6$, we only apply hard negative sampling, which is assumed to contribute to mAP performance.

\begin{table}[t!]
\small
    \setlength{\tabcolsep}{0.58\tabcolsep}
    \begin{center}
        \begin{tabular*}{\linewidth}{@{\extracolsep{\fill}} lccc }
            \toprule
            $f$ type & \# Params(M) & R@50 & mR@50 \\
            \midrule
            dot product attention & \textbf{41.3} & 25.9 & 6.2 \\
            dot product           & \textbf{41.3} & 27.4 & 6.8 \\
            Hadamard product      & 41.5 & 29.1 & 7.2 \\
            sum                   & 41.5 & 29.5 & 7.3 \\
            \textbf{concat}       & 41.6 & \textbf{29.9} & \textbf{7.9} \\
            \bottomrule
        \end{tabular*}
    \end{center}
    \caption{\textbf{Ablation for relation function.} Since concat represents the relationship between the attention query and attention key without loss of information, it shows the best performance among all $f$ variants.}
    \label{table:relation_function}
\end{table}

\begin{table}[t!]
\small
    \setlength{\tabcolsep}{0.58\tabcolsep}
    \begin{center}
        \begin{tabular*}{\linewidth}{@{\extracolsep{\fill}} cccc }
            \toprule
            $\mathcal{L}_\text{con}$ & connectivity score & R@50 & mR@50 \\
            \midrule
                       &            & 29.4 & 7.3 \\
            \checkmark &            &  29.4& 7.6  \\
            \textbf{\checkmark} & \textbf{\checkmark} & \textbf{30.2}& \textbf{7.9} \\
            \bottomrule
        \end{tabular*}
    \end{center}
    \caption{\textbf{Ablation for connectivity loss and score.} $\mathcal{L}_\text{con}$ denotes whether the model is trained with loss for connectivity prediction. \text{connectivity score} denotes whether we use connectivity score for sorting predicted relation triplets in evaluation. Results show that using connectivity loss and score both improve performance.}
    \label{table:connectivity_prediction}
\end{table}

\section{Additional Results}\label{sup:results}
\subsection{Ablation Studies}

\noindent
\textbf{Relation Function.}
We conduct an ablation study on the relation function $f$, as presented in~\cref{table:relation_function}. To evaluate the impact of different relation functions, we conduct experiments using only the self-attention relation sources $[R_a^1; ...; R_a^L]$ without the final relation source $R_z$. 
Furthermore, we do not use linear weights $W_S^l$ and $W_O^l$ for relation source representations; therefore, using the dot product attention function entails utilizing the self-attention weights in their original form.
Surprisingly, using only the attention weights of the object detector shows consistently high results, supporting our hypothesis that self-attention contains information relevant to relations. Additionally, excluding only the softmax function from dot product attention significantly improves performance. We also explore different element-wise functions for the relation function, including Hadamard product, sum, and concat. Among these, concat, which preserves the representations of attention query and key, exhibits the best performance.

\noindent
\textbf{Connectivity Prediction.}
We perform ablation studies on the connectivity loss $\mathcal{L}_\text{con}$ used during training and the connectivity score employed during inference.
As outlined in~\cref{table:connectivity_prediction}, both connectivity loss and connectivity score contribute to the performance improvements. 
These findings indicate that connectivity loss serves as a hint loss for the relation extraction loss, and the connectivity score effectively filters out candidates of object pairs that are less likely to have relations.

\subsection{Model Selection}

\textbf{Loss Function.} 
Due to the vast search space for the hyperparameters of the loss function, 
we first set the connectivity prediction loss coefficient $\lambda_\text{con}$ to $0$ and explore the relation extraction loss coefficient $\lambda_\text{rel}$ in increments of $5$, which is the bounding box L$1$ loss coefficient $\lambda_\text{L1}$, as shown in~\cref{table:lambda_rel}. Then we explore $\lambda_\text{con}$ in multiplies of tuned $\lambda_\text{rel}$ as shown in~\cref{table:lambda_con}. Since the relation tensor is sparse, relatively high loss coefficients improve the performance.

\begin{table}[t!]
\small
\centering
    \begin{center}
        \begin{tabular*}{\linewidth}{@{\extracolsep{\fill}} ccccc }
            \toprule
               $\lambda_\text{rel}$ & 5 & 10 & \textbf{15} & 20  \\
               \midrule
                R@$50$  & 29.1 & 29.1 & \textbf{29.4} & 28.8 \\
                mR@$50$ & 7.3  & 7.3  & \textbf{7.3}  & 7.0  \\
            \bottomrule
        \end{tabular*}
    \end{center}
    \caption{\textbf{Experiments for relation extraction loss coefficient $\lambda_\text{rel}$.} We fix $\lambda_\text{con}$ to $0$ and conduct experiments on $\lambda_\text{rel}$ first.}
    \label{table:lambda_rel}
\end{table}

\begin{table}[t!]
\small
\centering
    \begin{center}
        \begin{tabular*}{\linewidth}{@{\extracolsep{\fill}} ccccc }
            \toprule
               $\lambda_\text{con}$ & 0 & 15 & \textbf{30} & 45   \\
               \midrule
                R@$50$  & 29.4 & 29.7 & \textbf{30.2} & 29.6 \\
                mR@$50$ & 7.3  & 7.4  & \textbf{7.9}  & 7.4  \\
            \bottomrule
        \end{tabular*}
    \end{center}
    \caption{\textbf{Experiments for connectivity prediction loss coefficient $\lambda_\text{con}$.} $\lambda_\text{rel}$ is set to the optimal value $15$ from~\cref{table:lambda_rel}.}
    \label{table:lambda_con}
\end{table}

\noindent
\textbf{Adaptive Smoothing.}
To choose a hyperparameter $\alpha$ representing the minimum uncertainty for adaptive smoothing, we first set the hyperparameter range for $\alpha$. Since the uncertainty measured through the bipartite matching is sensitive to related configurations such as the number of object queries $N$ and weights used to calculate matching cost $\mathcal{L}_\text{match}$, we devise the method to explore the hyperparameter range in advance. We find the hyperparameter range by measuring the validation uncertainty when the model is initialized. To reflect the situation in which the model is initialized with random weights, the hyperparameter range is chosen so that the valid uncertainty can cover a wide range between $0$ and $1$.
We experiment with $\alpha$ of $10^{-13}$, $10^{-14}$, and $10^{-15}$ corresponding to valid uncertainties of $0.844$, $0.487$, and $0.135$, respectively. 
Results shown in~\cref{table:adaptive_smoothing} demonstrate that the performance is relatively robust regardless of the hyperparameters.
Judging from the fact that $10^{-14}$ where initial valid uncertainty is $0.487$ performs the best, setting initial valid uncertainty close to $0.5$ might be suitable for the situation where the model weights are randomly initialized.

\begin{table}[t!]
\small
    \setlength{\tabcolsep}{0.58\tabcolsep}
    \begin{center}
        \begin{tabular*}{\linewidth}{@{\extracolsep{\fill}} ccc }
            \toprule
            $\alpha$ & R@$50$ & mR@$50$ \\
            \midrule
            $10^{-13}$ & 30.1 & 7.8\\ 
            $\textbf{10}^{-\textbf{14}}$ & \textbf{30.2} & \textbf{7.9}\\
            $10^{-15}$ & 30.0 & 7.8\\
            \bottomrule
        \end{tabular*}
    \end{center}
    \caption{\textbf{Experiments for adaptive smoothing hyperparameter $\alpha$.} We set the hyperparameter range of $\alpha$ with the validation uncertainty of the initialized model. Hyperparameters within the range have similar performances; however, $10^{-14}$ shows the best performance.}
    \label{table:adaptive_smoothing}
\end{table}

\noindent
\textbf{Sampling Methodology.}
We explore hyperparameters $k_{\text{neg}}$ and $k_{\text{non}}$ for negative sampling and non-matching sampling, respectively. To narrow down the hyperparameter range, we set $k_{\text{neg}}$ equal to $k_{\text{non}}$. Results in~\cref{table:n_sample} illustrate a trade-off, where increasing the sampling coefficients enhances the amount of information on triplets not representing the ground truth relations, and decreasing the coefficient reduces the sparsity of the ground truth relation graph. We select $k_{\text{neg}}$ and $k_{\text{non}}$ as $80$, yielding the best R@$50$.

\begin{table}[t!]
\small
    \setlength{\tabcolsep}{0.58\tabcolsep}
    \begin{center}
        \begin{tabular*}{\linewidth}{@{\extracolsep{\fill}} cccc }
            \toprule
            $k_{\text{neg}}$ &$k_{\text{non}}$ & R@50 & mR@50 \\
            \midrule
            10 & 10 & 29.6 & \textbf{8.2} \\
            20 & 20 & 29.9 & \textbf{8.2}\\ 
            40 & 40 & 30.0 & 8.1\\
            \textbf{80} & \textbf{80} & \textbf{30.2} & 7.9\\
            160& 160 & 29.9 &7.7\\
            \bottomrule
        \end{tabular*}
    \end{center}
    \caption{\textbf{Experiments for sampling hyperparameter $k_{\text{neg}}$ and $k_{\text{non}}$.} 
    We choose $k_{\text{neg}}$ and $k_{\text{non}}$ as $80$ which shows the best R@$50$.}
    \label{table:n_sample}
\end{table}

In addition to sampling hyperparameters, we perform comprehensive experiments on sampling options, as presented in~\cref{table:sample_options}. For the negative region and non-matching region, we explore the following three options: one that considers the entire region without sampling (-), another that considers only a portion of it through sampling ($80$), and a third that does not consider the region ($0$). The results indicate that considering the entire region or ignoring it is suboptimal, and sampling in both regions is crucial for performance.

\begin{table}[t!]
\small
    \setlength{\tabcolsep}{0.58\tabcolsep}
    \begin{center}
        \begin{tabular*}{\linewidth}{@{\extracolsep{\fill}} cccc }
            \toprule
            $k_{\text{neg}}$ &$k_{\text{non}}$& R@$50$ & mR@$50$ \\
            \midrule
            0 & 80 & 29.4 & 7.4 \\
            \textbf{80} & \textbf{80} & \textbf{30.2} & \textbf{7.9}\\
            - & 80 & 30.0 & \textbf{7.9}\\
            \midrule
            80 & 0& 29.7 & 6.8\\
            \textbf{80} & \textbf{80} & \textbf{30.2} & \textbf{7.9}\\
            80 & - & 29.8 & 7.2\\
            \midrule
            - & 0 & 29.7 & 7.0\\
            - & - & 29.7 & 7.2\\
            0 & - & 29.2 & 6.8\\
            \bottomrule
        \end{tabular*}
    \end{center}
    \caption{\textbf{Experiments for sampling options.} ``-'' denotes that we use the whole region without sampling. $0$ indicates that the region is not considered. Sampling from both negative and non-matching regions shows the best performance.}
    \label{table:sample_options}
\end{table}

\subsection{Analysis}

\myparagraph{Backbone.}
To observe whether the performance improves with a heavier backbone, we conduct experiments using ResNet-$101$ as the backbone instead of ResNet-$50$. It shows improved object detection performance and relation extraction performance: AP50 $32.3$ (+$1.5$), R@$50$ $30.8$ (+$0.6$), and mR@$50$ $8.1$ (+$0.2$). 
However, the improvement in the relation extraction performance is relatively lower compared to the enhancement in the object detection performance.
We speculate that the capacity of the Transformer decoder might be more crucial than the CNN backbone for the performance of the relation extraction.

\begin{figure*}[t]
    \begin{center}
        \includegraphics[width=0.99\linewidth]{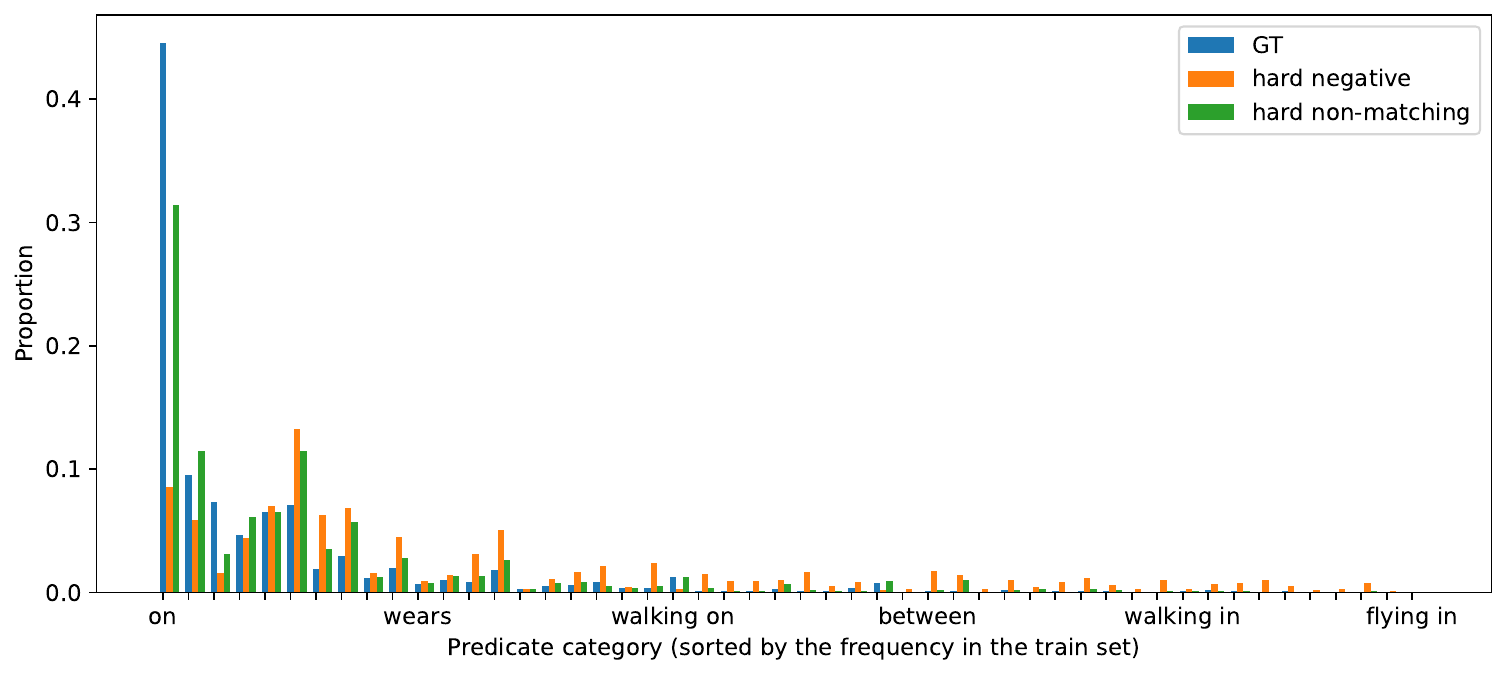}
    \end{center}
    \caption{\textbf{The comparison of predicate category distribution based on graph regions.} We compare the predicate categories of GT, hard negatives, and hard non-matchings for the validation dataset using histograms. We sort the predicate categories based on their frequency in the training dataset.}
    \label{fig:sampling_methodology}
\end{figure*}

\noindent
\textbf{Adaptive Smoothing.}
Since proposed adaptive smoothing can be applied to any one-stage SGG model that utilizes the explicit object detector, we apply the technique to Relationformer~\cite{shit2022relationformer} and SGTR~\cite{li2022sgtr}, where object detection loss is used and detected objects are related to relation extraction.
We conduct experiments using publicly available code and adapt the technique based on the characteristics of each model.
Since Relationformer uses softmax cross-entropy with an additional ``no relation'' class for the relation extraction, we apply smoothing for the ground truth relation class and compensate the target value of the ``no relation'' class by the same amount.
In our reproduced experiments, it shows improved performance: R@$50$ $26.61$($+0.12$), mR@$50$ $8.54$($+0.71$), and ng-R@$50$ $28.84$($+0.76$) where ng denotes no graph constraints.
Since SGTR matches detected objects with triplets through graph assembling, we apply relation smoothing on predicate labels based on the uncertainties of the detected objects matching the subjects and objects in triplets. Our smoothing method enhances overall performance: R@$50$ $24.36$($+0.04$) and mR@$50$ $12.88$($+0.75$).
The results demonstrate the generality of the adaptive smoothing.
Exploring the possibility of applying the adaptive smoothing based on matching costs of subjects and objects for triplet detection models that do not use an explicit object detector could be an interesting avenue for future research.

\noindent
\textbf{Sampling Methodology.}
We compare the distribution of predicate categories for hard negatives and hard non-matchings with that of the GT as shown in~\cref{fig:sampling_methodology}.
Since the non-matching region is composed of object candidates that do not match with the ground truth objects and object candidates that closely resemble ground truth objects are selected as hard non-matchings, hard non-matchings exhibit a prevalence of the head predicate categories similar to the GT.
On the contrary, hard negatives exhibit a relatively lower proportion of the head categories, and tail categories are more frequently selected.
As the negative region is constructed from object candidates that match the ground truth objects, it seems that hard negatives are selected from tail classes that are likely to exist in reality but are not annotated.

\begin{figure}[t!]
    \begin{center}
        \includegraphics[width=0.95\linewidth]{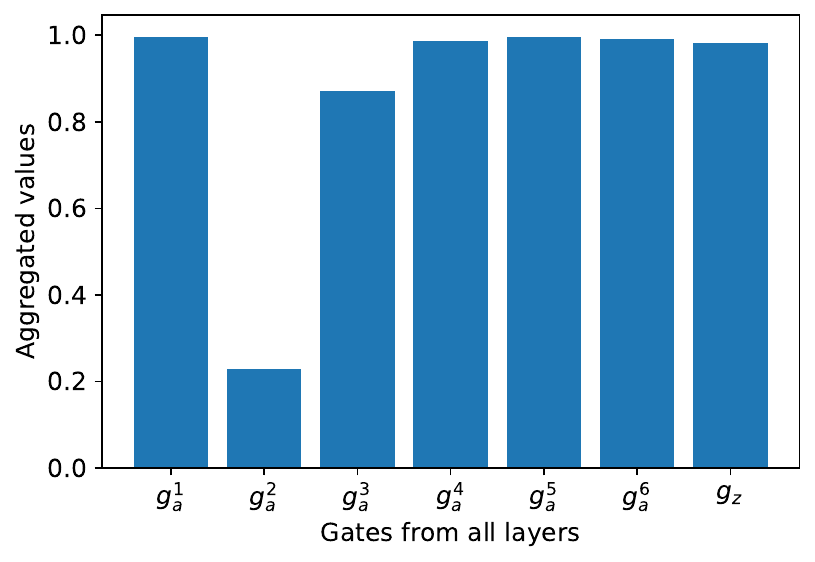}
    \end{center}
    \caption{\textbf{Aggregated gate values from all layers}. We aggregate the $N \times N$ shaped gate matrices for each layer and report the average over the entire validation dataset.}
    \label{fig:gated_sum}
\end{figure}

\noindent
\textbf{Gated Sum.}
We examine the utilization of relation source representations from each layer in the gating mechanism on~\cref{fig:gated_sum}.
Remarkably, the gate values for the first self-attention source representations, which precede the cross-attention layer and do not utilize any image information, are close to $1.0$.
Since object queries are trained to reflect the diverse distribution of objects in the training data~\cite{sun2021sparse}, the relationship information between object queries appears to be utilized as a primary bias, reflecting the various object relationships in the training data.
After passing through the cross-attention layer once, the gate values of the second self-attention source representations are very low. However, they gradually increase as the model incorporates image information well.
In particular, from the fourth self-attention source representations, the gate values are higher than those of the relation source representations in the final layer.

\begin{table}[t!]
\scriptsize
    \setlength{\tabcolsep}{0.58\tabcolsep}
    \begin{center}
        \begin{tabular*}{\linewidth}{@{\extracolsep{\fill}} lcccc }
            \toprule
            Model  & Backbone  & \# of params (M) & FPS & MACs (G) \\
            \midrule
            FCSGG~\cite{liu2021fully} & -  & 87.1 & 6.0 & 655.7 \\
            \midrule
            RelTR~\cite{cong2023reltr} & DETR-50  & \underline{63.7} & \underline{13.4} & \textbf{67.6} \\  
            \midrule
            SGTR~\cite{li2022sgtr}& DETR-101& 117.1 &  6.2 &  \underline{127.0} \\
            Iterative SGG~\cite{khandelwal2022iterative} & DETR-101  &  93.5 &6.0 & 130.3\\
            \midrule
            Relationformer~\cite{koner2020relation} & DDETR-50  & 92.9 & 8.5 & 336.7 \\
            \textbf{EGTR} (Ours) & DDETR-50  & \textbf{42.5}&\textbf{14.7 }& 132.4  \\
            \midrule
            SSR-CNN~\cite{teng2022structured}  & SRCNN-X101-FPN  & 274.3& 4.0 &  297.7 \\
            \bottomrule
        \end{tabular*}
    \end{center}
    \caption{\textbf{Efficiency of one-stage SGG models}. In addition to FPS, we measure MACs, which account for theoretical complexity. ``-50" represents ResNet50, ``-101" denotes ResNet-101, and ``-X101-FPN" signifies ResNeXt-101-FPN~\cite{xie2017aggregated}. ``DDETR" corresponds to Deformable DETR~\cite{zhu2020deformable}, and ``SRCNN" corresponds to Sparse-RCNN~\cite{sun2021sparse}. For a fair comparison, the image size is set to a minimum of $600$ for the shortest side and a maximum of $1000$ for the longest side. FPS is measured in a single V100.}
    \label{table:efficiency}
\end{table}

\subsection{Efficiency}

As depicted in~\cref{table:efficiency}, we report Multiply-ACcumulation (MACs) to assess efficiency in addition to the number of parameters and frames per second (FPS). MACs quantify the number of multiply and accumulate operations performed by a neural network during the inference phase. It is worth noting that MACs are estimated to be roughly half the number of Floating Point Operations (FLOPs).
For a fair comparison, the image size is set to a minimum of $600$ for the shortest side and a maximum of $1000$ for the longest side. 

It seems that EGTR has relatively high MACs, considering the superior efficiency in terms of the number of parameters and FPS.
However, it is noteworthy that our MACs are primarily attributed to the Deformable DETR backbone, and additional MACs from our relation extractor are only $16.8$G.
With $100$ object queries, Deformable DETR-$50$ shows $115.0$G MACs, compared to DETR-$50$ with $56.1$G.
Although Deformable DETR has more than twice the theoretical complexity compared to DETR, we opt for Deformable DETR due to its notably enhanced convergence speed~\cite{zhu2020deformable}.
Leveraging Deformable DETR as a backbone, we use a lightweight relation extractor composed of only $2.5$M parameters, resulting in the fastest inference speed.

\subsection{PredCls \& SGCls}
To assess how well the model can capture the structure of the scene given the ground truth objects information, we provide results for PredCls and SGCls.
As they were introduced in the two-stage SGG models to measure relation prediction given ground truth objects, measuring them in one-stage SGG models may involve some arbitrariness.
We reviewed one-stage studies~\cite{liu2021fully,cong2023reltr} that had reported PredCls and SGCls, and carefully designed measurements for one-stage SGG models.
We perform bipartite matching for object queries with ground truth objects and replace the prediction of the matched object queries with the corresponding ground truth objects' labels. 
Note that the representations of object queries used for the relation prediction remain unchanged.
As shown in~\cref{table:sgpredcls}, EGTR performs well in both PredCls and SGCls settings. 
It demonstrates that SGDet performance of EGTR does not solely depend on high object detection performance but also relation extraction performance.

\begin{table}[t!]
\small
\centering
    \begin{center}
        \begin{tabular*}{\linewidth}{@{\extracolsep{\fill}} ccccc }
            \toprule
               Models & AP$50$ & R@$50$   & mR@$50$  \\
               \midrule
               FCSGG~\cite{liu2021fully} & 28.5 & 41.0 / 23.5 / 21.3 & 6.3 / 3.7 / 3.6\\
               RelTR~\cite{cong2023reltr} & 26.4 &  \textit{36.0} / \textit{30.5}  / \textit{25.2} &  \textit{10.8} / \textit{9.3} /  \textit{\textbf{8.5}}\\ 
               \textbf{EGTR} (Ours) & \textbf{30.8} & \textbf{54.3}  / \textbf{39.8} / \textbf{30.2} & \textbf{16.6} / \textbf{11.9} /  7.9  \\
            \bottomrule
        \end{tabular*}
    \end{center}
    \vspace{-1.5em}
    \caption{\textbf{Comparison with one-stage SGG models on Visual Genome test set.} We report results for PredCls, SGCls, and SGDet settings, separated by ``/''. \textit{Italic} denotes that we remeasured the score with a publicly available model checkpoint for a fair comparison: the ground truth objects are utilized rather than ground truth triplets in the original RelTR report.}
    \label{table:sgpredcls}
\end{table}

\subsection{Zero-shot Performance}
We have noticed that popular technique frequency baseline~\cite{zellers2018neural} directly influences the zero-shot performance in our model. Without the frequency baseline, EGTR demonstrates a commendable zR@$50$ performance with a score of $2.1$ despite a decrease of $0.2$ points in R@$50$ and $0.6$ points in mR@$50$.

\begin{figure}[t!]
        \includegraphics[width=1\linewidth]{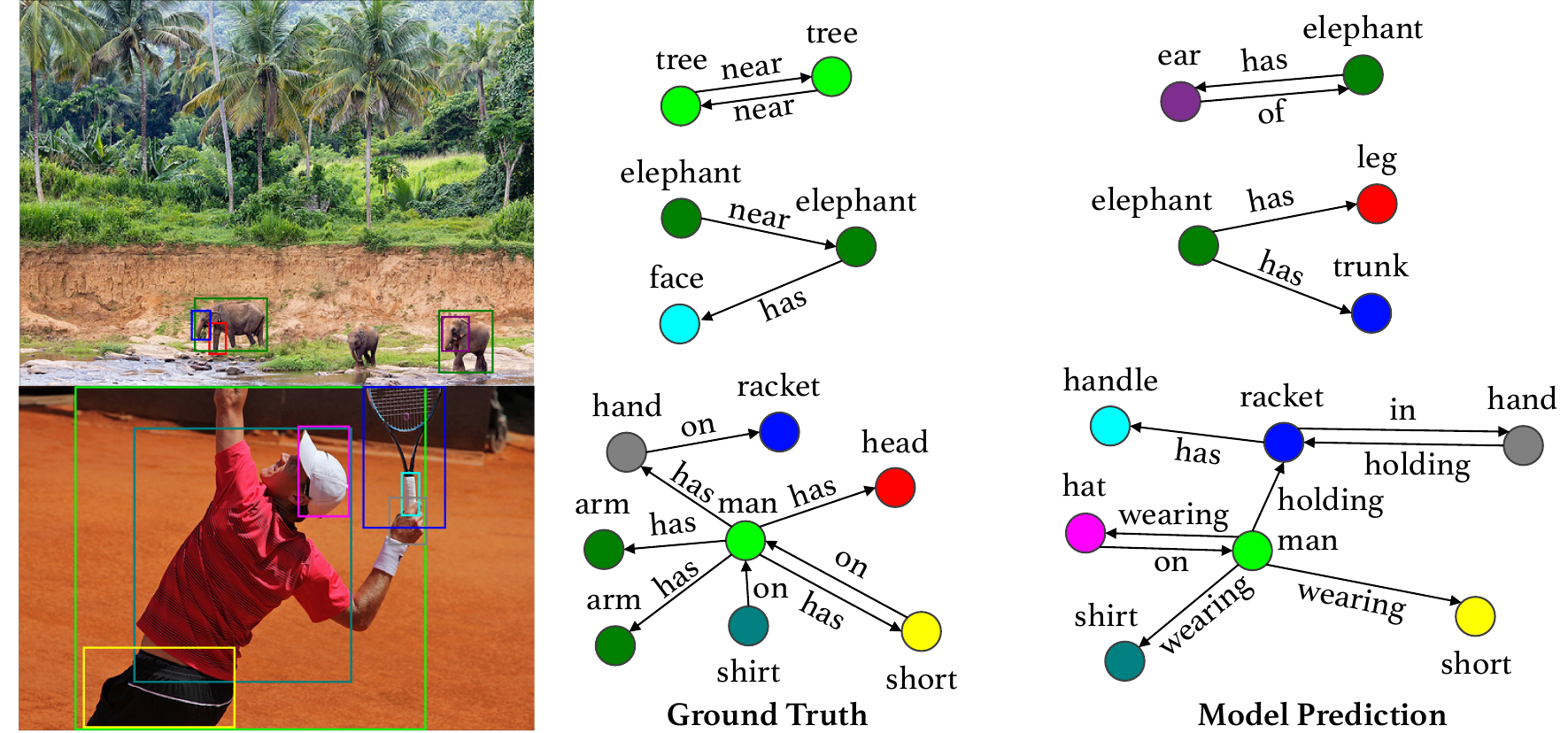}
    \caption{\textbf{Qualitative Analysis}. For visualization, we select the same number of predicted triplets as the ground truth triplets within the Visual Genome validation dataset.}
    \label{fig:quala}
\end{figure}

\subsection{Qualitative Results}
In~\cref{fig:quala}, we present qualitative examples of the Visual Genome validation dataset. The depicted results illustrate the capability of our methodology to generate relationships that are both plausible and semantically rich.

\clearpage